\documentclass{article}
\usepackage[final,nonatbib]{neurips_2021}

\usepackage[utf8]{inputenc} 
\usepackage[T1]{fontenc}    
\usepackage[colorlinks=true,citecolor=blue,linkcolor=red]{hyperref}

\usepackage{url}            
\usepackage{booktabs}       
\usepackage{amsfonts}       
\usepackage{nicefrac}       
\usepackage{microtype}      
\usepackage{xcolor}         

\usepackage{wrapfig,lipsum,booktabs}
\usepackage{graphicx}
\usepackage{subfig}

\usepackage{amsmath}
\usepackage{nccmath}
\usepackage{amssymb}
\usepackage{amsthm}
\usepackage{bbm}
\usepackage{multirow}
\usepackage[normalem]{ulem}
\usepackage{algorithm}
\usepackage{algorithmic}
\usepackage[para]{footmisc}

\newcommand{\beq}{ \begin{equation} }
\newcommand{\eeq}{ \end{equation} }
\newcommand{\E}{ \mathbb{E} }
\newcommand{\LDR}{{\sf{LDR}}}
\newcommand{\LDRM}{{\sf{LDRM}}}

\newcommand{\LDRV}{{\sf{LDRV}}}

\newcommand{\cD}{{\mathcal{D}}}
\newcommand{\cZ}{{\mathcal{Z}}}
\newcommand{\pd}{p_{\sf data}}

\newcommand{\var}{\text{var}}
\usepackage{makecell}
\usepackage{float}

\usepackage{url}
\usepackage{subfig}
\usepackage{comment}
\usepackage{xr-hyper} 
\title{Self-Diagnosing GAN: Diagnosing Underrepresented Samples in Generative Adversarial Networks}

\author{%
  Jinhee Lee\thanks{Equal contribution.}\\
  School of Electrical Engineering\\
  KAIST\\
  \texttt{jin.lee@kaist.ac.kr}
   \And
  Haeri Kim\footnotemark[1]\protect\phantom{\footnotesize 1}\thanks{This work was done as a student at KAIST.}\\
  Samsung Research\\
  Samsung Electronics\\ 
  \texttt{haeri.kim@kaist.ac.kr}
  \And
  Youngkyu Hong\footnotemark[1]\protect\phantom{\footnotesize 1}\footnotemark[2]\\
  NAVER AI Lab \\
  NAVER \\
  \texttt{youngkyu.hong@navercorp.com}
   \And
  Hye Won Chung\thanks{Corresponding author.}\\
  School of Electrical Engineering\\
  KAIST\\
  \texttt{hwchung@kaist.ac.kr}
}

\begin{document}

\maketitle

\begin{abstract}
Despite remarkable performance in producing realistic samples, Generative Adversarial Networks (GANs) often produce low-quality samples near low-density regions of the data manifold, e.g., samples of minor groups.
Many techniques have been developed to improve the quality of generated samples, either by post-processing generated samples or by pre-processing the empirical data distribution, but at the cost of reduced diversity. 
To promote diversity in sample generation without degrading the overall quality, we propose a simple yet effective method to diagnose and emphasize underrepresented samples during training of a GAN. The main idea is to use the statistics of the discrepancy between the data distribution and the model distribution at each data instance.  
Based on the observation that the underrepresented samples have a high average discrepancy or high variability in discrepancy, 
we propose a method to emphasize those samples during training of a GAN. 
Our experimental results demonstrate that the proposed method improves GAN performance on various datasets, and it is especially effective in improving the quality and diversity of sample generation for minor groups.
\end{abstract}

\section{Introduction}
\label{sec:intro}
Generative Adversarial Networks (GANs) have achieved remarkable performance in producing realistic samples for complex generation tasks, including image/video synthesis~\cite{brock2018large, mathieu2015deep}, style transfer~\cite{zhu2017unpaired, isola2017image}, and data augmentation~\cite{shrivastava2016training}.
However, GANs often fail to cover sparse regions of data manifold~\cite{stylegan2,instance}, leading to the underrepresentation of minor groups in the dataset~\cite{InclusiveGAN}.
In particular, GANs generate samples of minor groups with low fidelity or even fail to generate such samples, exhibiting the mode collapse~\cite{InclusiveGAN}. 

Many of previous techniques have focused on improving the overall sample quality of GANs, either by pre-processing the training dataset or by post-processing generated samples.
The pre-processing aims to remove instances that cannot be well-represented by GANs even before the training starts and gains fidelity on the focused samples~\cite{instance}.
A similar idea has been used to truncate the latent space by resampling or moving samples that fall outside of some acceptable range during training~\cite{stylegan2, biggan}. 
Post-processing, on the other hand, is a technique that can be applied after the training to remove low-quality generated samples by rejection sampling~\cite{DRS,turner2019metropolis}. 
All these approaches are effective in increasing the overall fidelity of samples from GANs, but reducing the diversity as a trade-off, and may exacerbate biases against the minor groups in sample generation.


In this work, we aim to improve diversity in sample generation without degrading the overall quality, with a special focus on coverage and quality improvement for minor groups.
Toward this, we design methods to detect and emphasize underrepresented samples in training of GANs.
Due to the 
lack of explicit labels available, detecting minor-subgroup samples is especially challenging for unsupervised learning.
Therefore, we first develop two new metrics, which can be easily calculated from a discriminator output of GANs, to detect underrepresented samples. 
The main idea is to measure the statistics (mean and variance) of the estimated discrepancy between the data distribution and model distribution at each data instance over multiple epochs of the training.  
The mean discrepancy indicates how close the data distribution is to the model distribution at each data over the training, while the variance in discrepancy measures how such discrepancy fluctuates across the training. We provide theoretical and empirical evidence that the mean discrepancy can effectively detect underrepresented samples, especially near collapsed modes, while the variance in discrepancy can detect minor data instances, which GANs suffer from modeling. 


Based on these observations, we propose a novel method to emphasize underrepresented samples during the training of GANs by score-based weighted sampling, where the score is defined as a weighted sum of the two metrics we devised. We validate our method with thorough experiments over controlled and real datasets and demonstrate the efficacy of the proposed sampling method in improving not only the overall quality (both fidelity and diversity combined) of sample generation but also the coverage and quality for semantic features of minor subgroups. 
Our contributions can be summarized as follows.

\begin{itemize}
\item We propose two new metrics, which can be simply computed from the discriminator, to diagnose GAN training and to detect underrepresented samples. 
By theoretical analysis and controlled experiments, we demonstrate that the proposed metrics are effective in detecting underrepresented minor samples. 
\item We propose an algorithm that can effectively emphasize underrepresented data by score-based weighted sampling during the training of GANs. 
Our experiments on controlled and real datasets show that our method improves diverse performance metrics on several GAN variants and enhances the coverage and quality of minor group generation.
\end{itemize}

Our code is publicly available at \url{https://github.com/grayhong/self-diagnosing-gan}.

\section{Related Work}
\label{sec:related}
\paragraph{Promoting data coverage in GANs}
Due to the unstable nature of the min-max game between a generator and a discriminator, GANs often suffer from mode collapse and produce samples with poor diversity. 
Several approaches have been proposed to promote better data coverage by modifying architectures~\cite{lin2020pacgan,unrollgan}, loss functions~\cite{arjovsky2017wasserstein,BWGAN} or adding regularizations~\cite{InfoGAN,berthelot2017began,Dist-GAN}. 
While effective in promoting overall data coverage, these approaches do not provide special care on minor modes and often fail to recover them when the minority ratio for certain feature is extremely low. We provide a method to promote data coverage for minor features even when the minority ratio is significantly low.

There exists another line of works to improve data coverage by designing hybrid generative models~\cite{srivastava2017veegan,rosca2017variational,InclusiveGAN}, which combines the idea of reconstructive models (e.g. variational autoencoder) to GANs, to take advantages of the reconstructive models in recovering diverse modes. 
This hybrid method, however, requires relatively high computational overhead to guarantee data coverage for all (or partial) real modes by optimizing reconstruction error in feature domain. 
Our method directly detects and emphasizes underrepresented samples so that the computational overhead is much lower.


\paragraph{Improving GAN performance by diagnosing samples} 
There have been promising attempts to improve GAN training by using the discriminator outputs to estimate the discrepancy between the data distribution and implicit model distribution. 
DRS~\cite{DRS} proposes the density ratio estimate based on the discriminator output to apply rejection sampling to filter generated samples. 
GOLD~\cite{GOLD} uses the similar estimate to re-weight fake samples to emphasize underrepresented fake samples. 
In \cite{ding2020subsampling} and \cite{biascorrection}, on the other hand, an external classifier is used to improve the density ratio estimates.  
There also exist some approaches to use discriminator outputs to select or weight ``useful'' fake samples during training.
Top-k training~\cite{sinha2020top} updates the generator by using only top-$k$ fake samples with the largest discriminator outputs.
In \cite{bridge} and \cite{improveGAN}, discriminator-based importance re-weighting schemes for fake samples are developed, 
and in \cite{wu2019logan}, latent samples are optimized to improve the fidelity. 

Our method uses the discrepancy estimate proposed in~\cite{DRS}, but its empirical mean and variance over multiple epochs, to extract more reliable and useful information to detect underrepresented minor group samples. 
We provide theoretical evidence of why not only the mean but also the variance of discrepancy estimate is effective in detecting underrepresented samples.
Our method detects and emphasizes underrepresented real samples, not the fake samples.
This difference is significant in promoting the data coverage of minor groups, since when fake samples already fail to cover minor modes, emphasizing a subset of fake samples cannot improve the data coverage for missed modes.

\section{Two Metrics to Detect Underrepresented Samples During GAN Training}
\label{sec:score}
\subsection{Measuring the discrepancy of GANs}\label{sec:background}

GAN training aims to train a generator with an implicit model distribution $p_g(x)$ that closely matches the data distribution $\pd(x)$. 
The discrepancy between $\pd(x)$ and $p_g(x)$ can be measured by the log density ratio $\log (\pd(x)/p_g(x))$, but it cannot be directly calculated in GANs, since $\pd(x)$ is unknown and $p_g(x)$ is implicit. 
Instead, the analysis in the original GAN paper~\cite{goodfellow2014generative} can be used to define an estimate on the density ratio by using the discriminator output as explained in~\cite{DRS}. 

The original GAN solves the min-max optimization $\min_G\max_D V(D,G)$ for the loss $V(D,G)=\E_{x\sim \pd}[ \log D(x)]+\E_{z\sim p_z} [\log(1-D(G(z)))]$. 
For any fixed generator $G$, the optimal discriminator yields $D^*(x)=\frac{\pd(x)}{\pd(x)+p_g(x)}$ and this allows us to define the Log-Density-Ratio estimate (LDR) by
\beq\label{eqn:LDR}
\LDR(x):=\log({D(x)}/({1-D(x)})).
\eeq
When $D(x)=D^{*}(x)$, the $\LDR(x)$ is equal to the log density ratio $\log (\pd(x)/p_g(x))$. 
When $\LDR(x)> 0$, the data point $x$ is underrepresented in the model, i.e., $\pd(x)>p_g(x)$, while when $\LDR(x)< 0$, the data is overrepresented, i.e., $\pd(x)<p_g(x)$.
Thus, we can leverage the value of $\LDR(x)$ of each instance $x$ to give feedback to improve the generator if the estimation is valid.

Some prior works have used the LDR estimate to improve GAN training.
As an example, GOLD \cite{GOLD} uses $\LDR(x)$ to evaluate the quality of the fake samples and re-weights the underrepresented fake samples when training the generator for conditional GANs.
However, we later show that re-weighting fake samples is less effective  than re-weighting real samples in improving diversity in sample generation.
We also empirically show that $\LDR(x)$ is an unstable metric to use. 
More detailed arguments are available in the Appendix~\S\ref{supp:LDR}.

As a remedy, we propose to use statistics of $\LDR(x)$, which are much more stable and informative metrics, to detect underrepresented data regions during the training. The main intuition is to use training dynamics--the behavior of a model as training progresses--to diagnose the learning behavior of each sample. 
In supervised learning, training dynamics have been widely studied to detect ``hard-to-learn'' samples~\cite{chang2017active, carto, wu2020curricula}. However, in learning generative models, the metrics to diagnose training dynamics are not clear since there is no explicit reference to measure the accuracy of the model. 
Here we define metrics that estimate the mean and variance of the discrepancy of GANs, LDRM (LDR Mean) and LDRV (LDR Variance), at each sample $x$ across the training steps $T = \{t_s, ..., t_e\}$: 
\begin{equation}
    \LDRM(x;T) = \frac{1}{|T|}\sum_{k \in T} \LDR(x)_k, \quad
         \LDRV(x;T)  = \frac{1}{|T|-1}\sum_{k \in T} \left[\LDR(x)_k - \LDRM(x;T)\right]^2\label{eqn:LDRMLDRV},
\end{equation}
where $\LDR(x)_k$ is the recorded LDR estimate~\eqref{eqn:LDR} in the $k$-th training step.
$\LDRM(x)$ measures how close $\pd(x)$ is to $p_g(x)$ over the training at sample point $x$, while $\LDRV(x)$ measures how such discrepancy fluctuates across training. 

Intuitively, samples that have been well-learned and generalized will have consistently small $\LDR(x)$ since $D(x)\approx 1/2$ (i.e., $\pd(x)\approx p_g(x)$), thus will exhibit low LDRM and LDRV, while underrepresented ``hard-to-learn'' samples will show high LDRM or LDRV values.
In the rest of this section, we thoroughly study the characteristics of data instances with high LDRM or high LDRV. 

\subsection{LDRV is effective in detecting samples from minor groups}\label{subsec:ldrv}

\begin{figure*}[t]
\centering
\begin{tabular}{ccc}
	\subfloat[Major\label{fig:color_mnist_major}]{\includegraphics[width=0.11\linewidth]{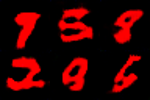}} & 
	\subfloat[Minor\label{fig:color_mnist_minor}]{\includegraphics[width=0.11\linewidth]{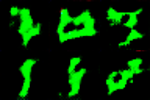}} &
	\\
	\subfloat[Major\label{fig:fmnist_major}]{\includegraphics[width=0.11\linewidth]{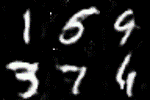}} & 
	\subfloat[Minor\label{fig:fmnist_minor}]{\includegraphics[width=0.11\linewidth]{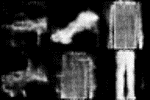}} &
    \multirow{-8.7}{*}{\subfloat[Partial Recall of major/minor groups vs. minority level \label{fig:recall_plot}]{\includegraphics[width=0.52\linewidth, height=0.22\linewidth]{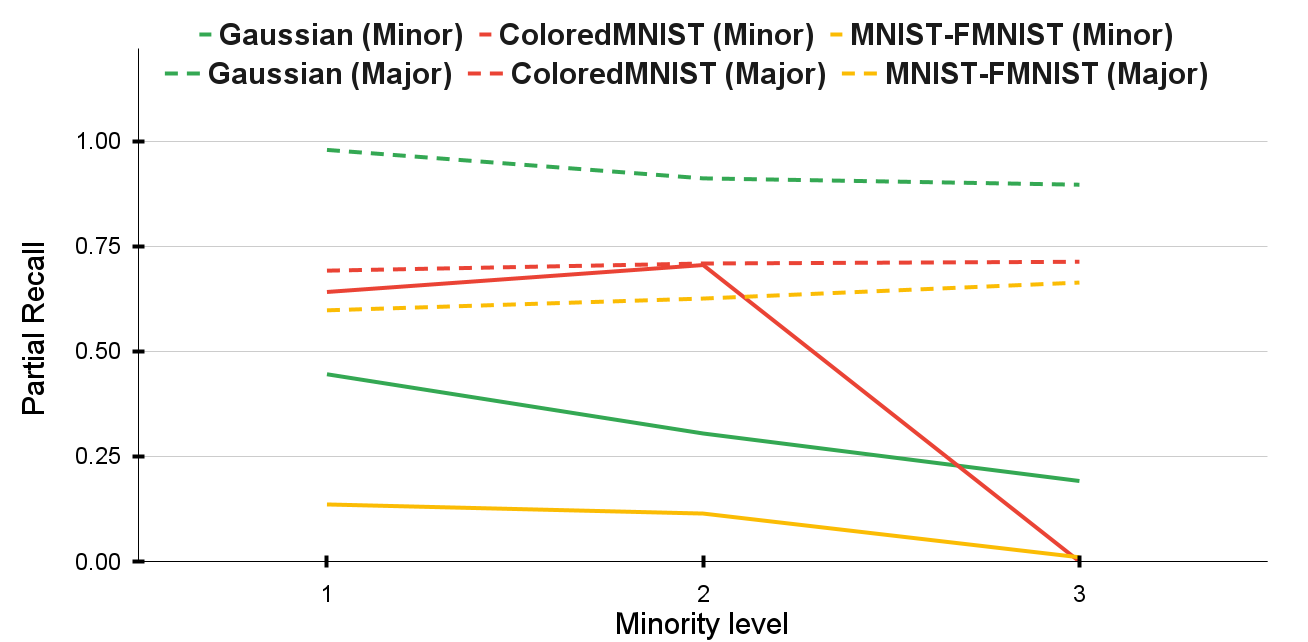}}}\\
    \end{tabular}
    \caption{Analysis on generated samples of GAN trained with (1) Single-mode Gaussian, (2) A mixture of MNIST (major) and FMNIST (minor), and (3) Colored MNIST with Red (major) and Green (minor) samples. (a) $\sim$ (d) show the examples of generated samples with major/minor features, and (e) shows the Partial Recall of major (dotted)/minor (solid) samples in each dataset on various minority levels. Both the sample quality and partial recall rate are higher for major groups.}
    \label{fig:minor_recall}
\end{figure*}
\paragraph{GANs have poor modeling for minor samples}
GANs are known to struggle with modeling minor samples~\cite{stylegan2}.
To scrutinize this phenomenon, we use following toy datasets each of which includes major and minor group: (1) Single-mode Gaussian with distance from the origin as a factor dividing two groups, (2) A mixture of MNIST (major) and FMNIST (minor), and (3) Colored MNIST with Red (major) and Green (minor) digits.
We vary the size of the minor group and define a \textit{minority level} to represent the scarcity of the minor group, i.e., a higher level indicates the scarcer minor group. 
Details of each dataset are available in the Appendix~\S\ref{supp:simul}.
Figure~\ref{fig:minor_recall} shows the poor quality of generated samples with minor features, relative to major features.
To quantify the level of underrepresentedness, we examine the coverage of modes for major vs. minor groups with the Partial Recall~\cite{PR}, which is the portion of the subset of real samples that reside in the manifold of the fake samples. 
As shown in Figure~\ref{fig:recall_plot}, major and minor groups have large recall gap and the gap gets worsen as the minority level gets severe.
This observation indicates that the minor group suffers not only the poor quality problem but also the low coverage problem, and it gives a strong motivation to detect the minor samples and emphasize them.

\paragraph{LDRV and minor samples}
We next provide heuristic arguments that LDRV can be used to detect samples with minor features, i.e., features of minor groups.
In particular, we show that minor samples tend to have higher LDRV values.
First, we view the discriminator as the logistic regression model: for each input $x_i$ the discriminator takes the inner product between the feature vector  $\phi_i=F(x_i)$ and the weight vector $\theta$ of the last layer to produce the reality score (the probability that the sample $x_i$ is real ($y_i=1$)), i.e.,
\beq\label{eqn:D_likelihood}
D({x}_i;\theta)={1}/{(1+e^{-{\theta}^T \phi_i })}=p(y_i=1|\phi_i,\theta).
\eeq
From a Bayesian perspective, assuming that prior distribution of $\theta$ is $p(\theta)=\mathcal{N}(\theta|0, s_0 I)$, the posterior distribution over $\theta$ is given by
$
p(\theta|(\phi_i,y_i)_{i=1}^n)\propto p(\theta)p(y_1^n|\phi_1^n, \theta).
$
To obtain a Gaussian approximation to the posterior distribution, we first find the maximum a posteriori estimate $\theta_{\sf MAP}$ that maximizes $\log p(\theta|(\phi_i,y_i)_{i=1}^n)$, which defines the mean of Gaussian. The covariance is then given by the inverse of the matrix of second derivatives of the negative log likelihood, which takes the form
\beq\label{eqn:sn_approx}
S_n = \left(\sum_{i=1}^n D({x}_i;\theta)(1-D({x}_i;\theta)) \phi_i \phi_i^T +\frac{1}{s_0}I\right)^{-1}.
\eeq
Lastly, approximating $D({x}_i;\theta)$ in~\eqref{eqn:D_likelihood} by the Taylor expansion at $\theta=\theta_{\sf MAP}$, LDRV can be expressed as
\beq
    \label{eqn:ldrv_approx}
\LDRV(x_i)\approx \text{var}\left(\log({D({x}_i;\theta)}/({1-D({x}_i;\theta)}))\right) \approx \phi_i^TS_n\phi_i.
\eeq
Details of the analysis is available in the Appendix~\S\ref{supp:analysis}.
\begin{table}[b]
\centering
\caption{Averaged LDRV of major/minor groups on various datasets with majority rate 90\%.}
\label{tab:minor_ldrv}
\vspace{0.2em}
\begin{tabular}{c|ccc}
\toprule
Group &Gaussian ($\sigma$=3.0)  & Colored MNIST & MNIST-FMNIST \\
\midrule
Major & 0.001 & 0.077 & 0.082 \\
Minor & 0.098 & 0.186 & 0.115 \\
\bottomrule
\end{tabular}
\end{table}

This analysis shows an important aspect regarding LDRV and minor features.
First, ~\eqref{eqn:ldrv_approx} shows that as the feature vector $\phi_i$ becomes more correlated with the principal components of $S_n$ (eigenvectors with largest eigenvalues), its $\LDRV$ gets larger.
Since each eigenvalue of $S_n$ is the reciprocal of that of $S_n^{-1}$, we consider the characteristics of the eigenvector $v$ of $S_n^{-1}$ with the least eigenvalue, which is the minimizer of
\beq
\begin{split}\label{eqn:inner}
\left\langle y, S_n^{-1} y\right\rangle= \sum_{i=1}^n D({x}_i;\theta)(1-D({x}_i;\theta)) \langle  y,\phi_i\rangle^2+\text{const}.
\end{split}
\eeq
Eq.~\eqref{eqn:inner} shows if $y$ does not align with (or orthogonal to) majority of feature vectors $\{\phi_i\}$ having $D(1-D)>0$, then it tends to have a smaller eigenvalue. 
Since a minor feature vector $\phi_j$ may have a small component on the eigenspace formed by the majority of $\{\phi_i\}$ having $D(1-D)>0$, when we plug in $y=\phi_j$ into~\eqref{eqn:inner}, the summation becomes small. This shows that the minor feature vector $\phi_j$ is correlated with the least eigenvector $v$ of $S_n^{-1}$ and thus it will have higher LDRV.

In Table~\ref{tab:minor_ldrv}, we show that minor group indeed has higher LDRV. Thus, both theoretical and empirical evidence shows that we can detect minor samples by investigating LDRV of training samples.
\subsection{LDRM is effective in detecting missing modes}
\begin{figure*}[!tb]
\centering
	\subfloat[LDRM and generated samples\label{fig:25_gaussain_ldrm}]{\includegraphics[width=0.35\linewidth]{{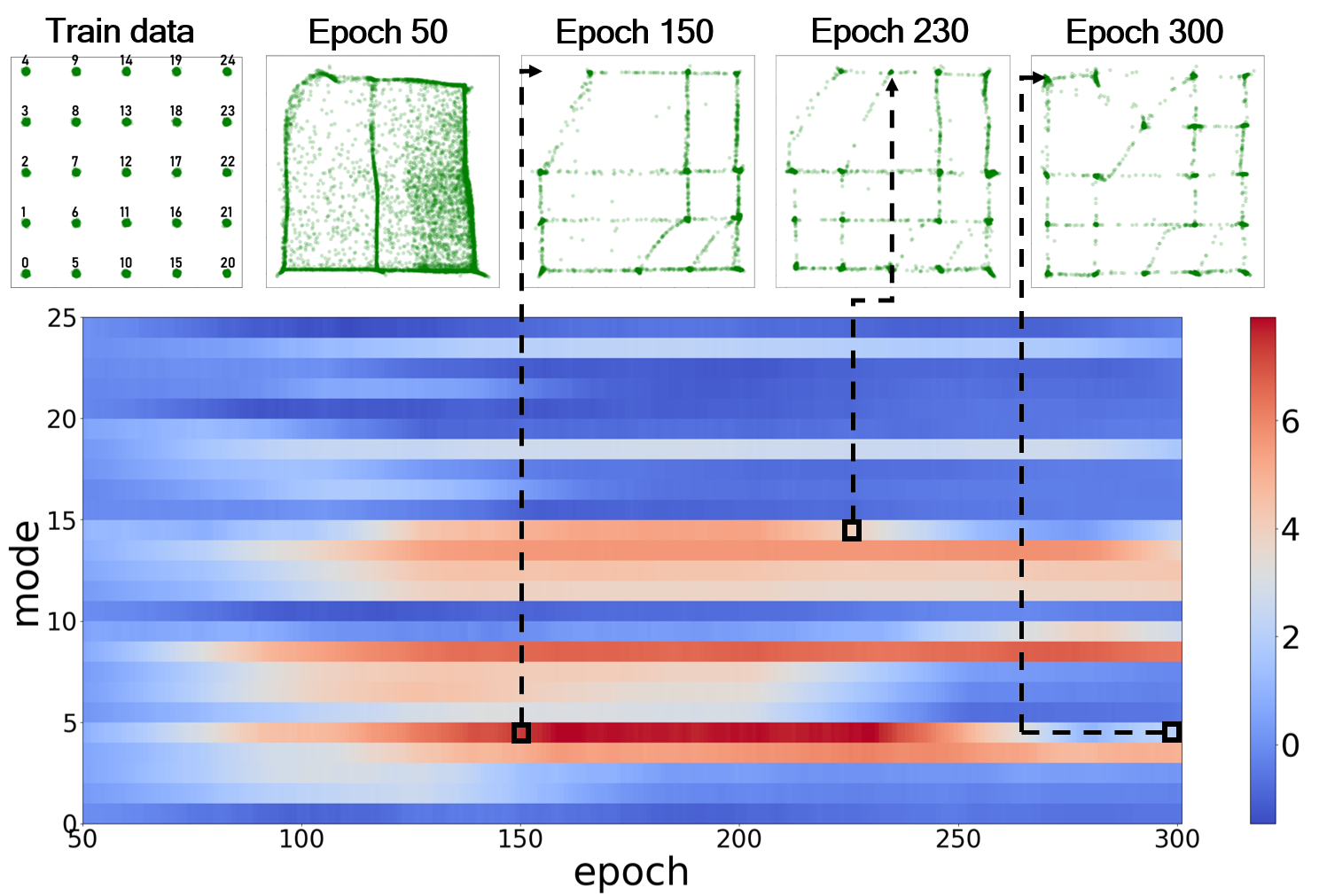}}} \qquad
    \subfloat[LDRM distribution and the number of high-quality samples over modes  \label{fig:25_gaussain_mode}]{\includegraphics[width=0.35\linewidth]{{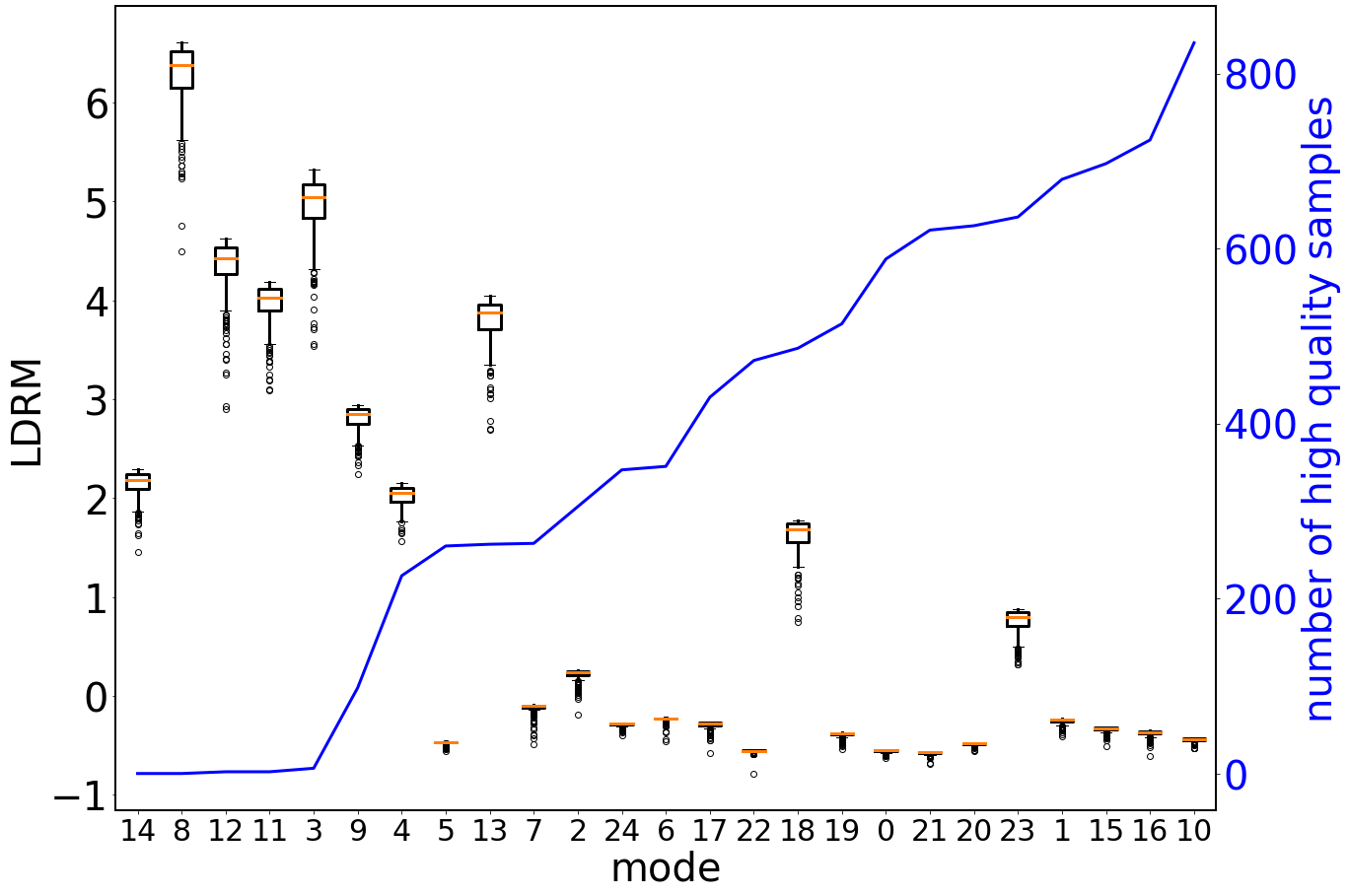}}}
    \caption{(a) Training dynamics of 25 Gaussians. The index of each mode is equal to $5x+y$ where the coordinates $(x,y)\in \{0,1,2,3,4\}^2$.  LDRM of training samples is recorded for each mode  with window $|T|=50$. Modes with high LDRM do not appear in the generated samples. (b) The empirical distribution of LDRM  (box plot) for each mode from  training samples, compared with the number of high-quality generated samples from each mode (blue).  LDRM can effectively detect dropped modes.}
    \label{fig:25_gaussian}
\end{figure*}

\paragraph{Mixture of 25 Gaussians}
From the definition of LDRM~\eqref{eqn:LDRMLDRV}, high LDRM samples $x$ tend to have smaller $p_g(x)$ than $\pd(x)$ over the training, thus are underrepresented. 
We next investigate the ability of LDRM to detect the regions of data manifold not yet covered by the model distribution $p_g(x)$.
We consider a mixture of 25 2D isotropic Gaussian distributions \cite{lin2020pacgan, yu2018generative, turner2019metropolis, DRS}.
During training, we record $\LDR(x)$ of the training samples and calculate LDRM values with window size $|T|=50$. 
We inspect LDRM values averaged over samples of each mode during the training.
As shown in Fig.~\ref{fig:25_gaussain_ldrm}, we observe that samples from underrepresented modes have higher mean LDRM values. 
This implies that we can detect the mode recovery by inspecting the mean of LDRM values. 

To further examine the mode recovery in generated samples, 
we assign each generated sample to its closest mode and consider it as a ``high-quality'' sample if it is within four standard deviations from its assigned mode~\cite{yu2018generative, DRS}. 
We then count the number of high-quality samples of each mode among 10,000 generated samples and analyze the correlation between the high-quality sample counts and the distribution of LDRM.
As shown in Fig. \ref{fig:25_gaussain_mode}, modes with only a few high-quality samples tend to have higher LDRM. 
This indicates that LDRM of the data instances can be used to detect the regions of data manifold not yet covered by the model, even without looking at the generated samples.

\section{Algorithm to Emphasize Underrepresented Samples}
\label{sec:algorithm}
\subsection{Proposed method: Stochastic Gradient Descent (SGD) sampled by discrepancy}\label{sec:alg}

We propose a simple modification to the GAN training procedures by using score-based weighted sampling for mini-batch SGD to emphasize underrepresented samples.
Let $\cD=\{x_i\}$ be the training dataset. 
The mini-batch of size $B$ for the training dataset is formed by $\cD_B=\{x^{(j)}: x^{(j)}=x_i \text{ where } i\sim P_s(i) \text{ for } j=1,\dots, B\}$, i.e., each sample $x_i\in \cD$ is sampled with certain probability $P_s(i)$. 
Our objective is to design the sampling frequency $P_s(i)$ that can emphasize underrepresented samples.
Based on the observations in Section~\ref{sec:score}, we first devise the discrepancy score $s(x_i;T)$ that reflects the underrepresentedness of each sample as follows: 
\beq \label{eqn:discrepancy}
	s(x_i;T) = \LDRM(x_i;T) + k \sqrt{\LDRV(x_i;T)},
\eeq
where $T$ is the set of steps used to calculate the discrepancy scores and $k$ is the hyperparameter to modulate the contribution of each statistic. 
The score~\eqref{eqn:discrepancy} can be interpreted as an upper limit of the confidence interval of LDR estimate, or weighted sum of LDRM and the square root of LDRV with weight controlled by $k$.
To ensure every data is sampled with at least some chance, we clip the minimum value of $s(x_i)$ to be $\epsilon=0.01$ (\texttt{min\_clip}) and clip the maximum value to have max-min ratio of 50, i.e., ${\max s(x)}/ {\min s(x)}=50$ (\texttt{max\_clip}).
For the clipped score $s'(x_i;T) = \texttt{max\_clip}(\texttt{min\_clip}(s(x_i;T)))$, our final weighted sampling frequency is
$
	P_s(i) = \frac{s'(x_i;T)}{\sum_ {j=1}^{|\cD|} s'(x_j;T)}.\label{eqn:discrepancy_P}
$
\subsection{Sample analysis of the discrepancy score}
\begin{figure*}[!tb]
\centering
\begin{tabular}{cc}
	\subfloat[Images with lowest disc. score \label{fig:cifar10_small}]{\includegraphics[width=0.32\linewidth]{{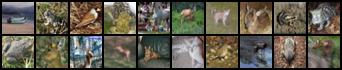}}} &\\

    \subfloat[Images with highest disc. score \label{fig:cifar10_large}]{\includegraphics[width=0.32\linewidth]{{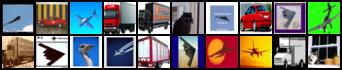}}}&\\

    \subfloat[Generated samples\label{fig:cifar10_gen}]{\includegraphics[width=0.32\linewidth]{{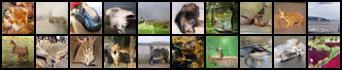}}}&  \multirow{-12.8}{*}{    \subfloat[Histogram of pixel count over intensity level    \label{fig:cifar10_histogram}]{\includegraphics[width=0.44\linewidth,height=0.32\linewidth]{{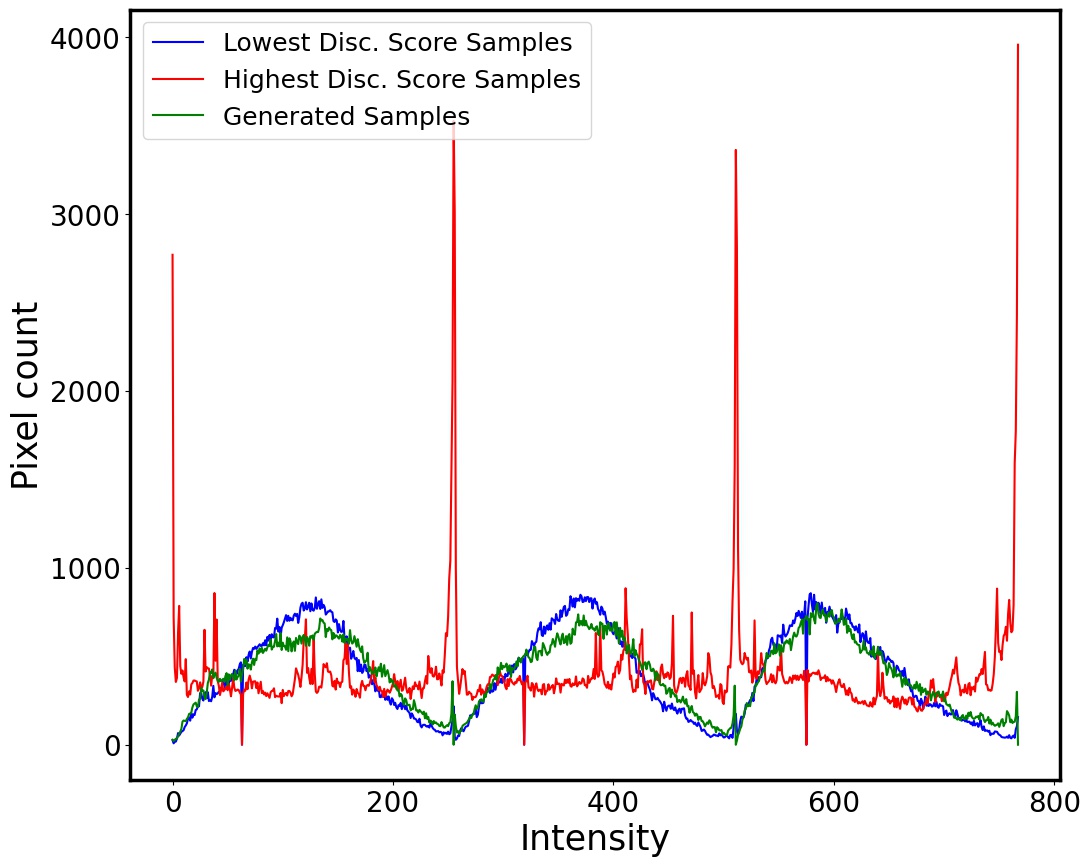}}}}\\
    \end{tabular}
    \caption{Training images with (a) lowest, (b) highest discrepancy scores, and (c) generated samples. Generated samples resemble training images with lowest score. (d) A smoothed histogram of the intensities for 100 samples per group.
    The intensity levels of RGB channels are concatenated, resulting in total  $768=256\times3$ levels. 
    Images with the lowest scores (blue) and generated samples (green) have a similar distribution, while images with the highest scores (red) show a high discrepancy. }
    \label{fig:cifar10_samples}
\end{figure*}

To check whether our discrepancy score indeed captures the underrepresented samples, we analyze the samples with lowest/highest discrepancy scores.
We train SNGAN~\cite{sngan} on CIFAR-10~\cite{CIFAR} for 40k steps and measure the discrepancy score of each sample.
We first present the images with lowest (Fig.~\ref{fig:cifar10_small})/highest (Fig.~\ref{fig:cifar10_large}) discrepancy scores among training images, and compare them with generated samples (Fig.~\ref{fig:cifar10_gen}).
High-scoring images have properties that are distinct from the generated samples (e.g., unusual background or shape), while low-scoring images contain features that are also available in generated samples.
Comparing the pixel intensity histogram (Fig.~\ref{fig:cifar10_histogram}) reveals the difference more clearly in sample properties.
Images with lowest discrepancy scores exhibit similar intensity distribution with generated samples, while images with highest scores appear to show an extremely different tendency.
We also analyze the Partial FID (FID~\cite{heusel2017gans} measured with a subset of training samples)  of lowest/highest-score groups.
The highest-score group has a Partial FID of 94.64 while the lowest-score group has 22.43.
The large gap between the two groups states that the generator fails to generate samples similar to high-score group.
These results imply that our discrepancy score successfully identifies underrepresented data that may need emphasis in further training.
For more examples of images with scores both for CIFAR-10 and CelebA, see the Appendix~\S\ref{supp:images_DS}.

\subsection{Post-processing by discriminator rejection sampling with auxiliary discriminator}\label{sec:algDSR}

Our weighted sampling gives bias toward underrepresented samples during training.
Though effective in improving diversity, this results in modified data distribution $\pd'(x) = f(x) \pd(x)$ where $f(\cdot)$ is the normalized sampling frequency.
Thus, the trained model distribution $p_g$ may be different from the original data distribution $\pd$. 
To solve this, we utilize the Discriminator Rejection Sampling (DRS)~\cite{DRS} to correct the bias after training. 
The rejection sampling  accepts a generated sample with probability $\frac{\pd(x)}{Mp_g(x)}$ for some constant $M>0$. 
To conduct rejection sampling, DRS method needs an estimate for ${\pd(x)}/{p_g(x)}$ calculated based on the discriminator outputs.  Since our discriminator is trained with biased $\pd'(x)$, we add an auxiliary discriminator  and train it with uniform sampling (i.e., without applying our sampling technique) during the weighted sampling procedure to obtain the LDR estimate  \eqref{eqn:LDR} for DRS, and use this measure for the rejection sampling of generated samples.

\subsection{Self-Diagnosing GAN (Dia-GAN) }\label{sec:diagan}
The overall algorithm (with details in the Appendix~\S\ref{supp:alg}) can be summarized as below:\\
\noindent\textbf{Phase 1 - Train and Diagnose:} Train GAN and evaluate the discrepancy score for each data instance. \\
\noindent\textbf{Phase 2 - Score-Based Weighted Sampling:} Encourage GAN to learn underrepresented regions of data manifold through score-based weighted sampling (Section~\ref{sec:alg}).\\
\noindent\textbf{Phase 3 - DRS:} After GAN training, correct the model distribution $p_g(x)$ by rejection sampling.

\section{Experiments}
\label{sec:experiments}
\subsection{Evaluation metrics and baselines}
\label{sec:5.1_baselines}
\paragraph{Evaluation metrics}
To evaluate the effect of our method on learned model distribution, we use various performance metrics including (1) Fr\'echet Inception Distance (FID)~\cite{heusel2017gans}, (2) Inception Score (IS)~\cite{salimans2016improved}, and (3) Precision and Recall (P\&R)~\cite{PR}. 
In addition to these global evaluation metrics, we consider (4) Reconstruction Error (RE). 
RE score is calculated by first training a convolutional autoencoder (CAE) with generated samples, and then calculating Euclidean distance between each training data and its reconstruction. 
RE can assess whether $p_g(x)$ covers $\pd(x)$ since CAE is known to have high RE for out-of-distribution samples~\cite{anomaly_drae,anomaly_rdae}. 
For more details, see the Appendix~\S\ref{supp:eval}.

\paragraph{Baselines} We compare the effect of our method with other methods that use the discriminator output for improving GAN training; 1) DRS~\cite{DRS}, 2) Gap of log-densities (GOLD)~\cite{GOLD}, and 3) Top-k training~\cite{sinha2020top}. 
GOLD\footnote{As the original GOLD estimator is designed for conditional GANs~\cite{cGAN}, we consider the unconditional version by removing the conditional discrepancy term. } uses the LDR estimate on generated samples to re-weight underrepresented samples (having high LDR) during training of GANs.
Top-k training uses only top-$k$ fake samples with the largest discriminator outputs, i.e., the samples believed to be the ``most realistic'', during the training of the generator. 
As our algorithm uses DRS after the training, we also analyze each method's performance with post-processing by DRS to measure the exact gain from our sampling method.

\subsection{GAN performance enhancement on real datasets}\label{sec:exp_real}
\begin{table}[t]
\centering
\caption{Comparison of diverse sampling/weighting methods for CIFAR-10/CelebA image generation.}
\label{table:cifar10-celeba-results}
\vspace{0.2em}
\begin{tabular}{c|cc|cc|ccc|ccc}
\toprule
Dataset & \multicolumn{4}{c|}{CIFAR-10} & \multicolumn{6}{c}{CelebA}\\
\midrule
\multirow{2.5}{*}{Methods} &\multicolumn{2}{c|}{SNGAN}&\multicolumn{2}{c|}{SSGAN}&\multicolumn{3}{c|}{SNGAN}&\multicolumn{3}{c}{SSGAN}\\
\cmidrule{2-11}
& FID $\downarrow$    &IS $\uparrow$  & FID $\downarrow$    & IS $\uparrow$ & FID $\downarrow$  & P $\uparrow$ & R $\uparrow$ & FID $\downarrow$  & P $\uparrow$& R $\uparrow$ \\
\midrule
Vanilla & 26.90 & 7.36 & 22.01 & 7.65& 7.12 & 0.68  & 0.44& 7.19 & 0.68 & 0.44\\
DRS~\cite{DRS} & 24.54 & 7.57 & 20.51 & 7.77 & 7.04 & 0.68 & 0.44& 7.08 & 0.68 & 0.45\\
GOLD~\cite{GOLD} & 28.86 & 7.21 & 21.90 & 7.57 & 7.31 & \textbf{0.69} & 0.44& 7.46 & 0.68 & 0.43\\
GOLD + DRS & 24.65 & 7.53 & 19.36 & 7.79 & 6.97 & 0.68 & 0.44 & 7.15 & 0.67 & 0.45\\
Top-k~\cite{sinha2020top} &  24.45 &  7.60 & 20.01 & 7.78  & 7.35 & 0.67 & 0.44& 7.23 & 0.67 & 0.45\\
Top-k + DRS &  23.92 &  7.70 & 20.09 & 7.88 & 7.35 & 0.68 & 0.44 & 7.16 & \textbf{0.68} & 0.45\\
\textbf{Dia-GAN} & \textbf{19.66} & \textbf{7.95}& \textbf{16.31} & \textbf{8.14} & \textbf{6.70} & 0.64 & \textbf{0.48}& \textbf{6.88} & 0.66 & \textbf{0.46}\\
\bottomrule
\end{tabular}
\end{table}

\paragraph{Experiments on CIFAR-10 and CelebA}
We first assess our method on two widely-studied GAN benchmark datasets, CIFAR-10~\cite{CIFAR} and CelebA~\cite{CelebA}. 
We evaluate our method on state-of-the-art GANs; SNGAN~\cite{sngan} 
and SSGAN~\cite{ssgan} with non-saturating variant of the original loss.
We train our model for 50k (75k) steps for CIFAR-10 (CelebA), where for our method and GOLD, the phase 1 takes 40k (60k) steps, and the phase 2 takes the remaining. 
We record LDR every 100 steps and use the last 50 records for calculating the discrepancy score.
For the discrepancy score~\eqref{eqn:discrepancy}, we use $k=0.3$ $(5.0)$ for CIFAR-10 (CelebA). 
Detailed configurations and hyperparameter search procedure are available in the Appendix~\S\ref{supp:simul}.

In Table~\ref{table:cifar10-celeba-results}, we first compare FID and IS over various methods on the CIFAR-10 dataset. 
Our proposed Dia-GAN achieves the best FID and IS with a great margin among all baseline methods in every GAN variant.
This result demonstrates the wide applicability and effectiveness of our method in improving the overall quality (fidelity and diversity combined) of generated samples.
Moreover, the comparison between DRS and our method assures that most of the gain indeed comes from our resampling method.
Also, we compare FID and P\&R over the methods on CelebA.
Our method consistently improves FID over baseline GANs.
Precision \& Recall analysis shows more detailed reasons for the improvement of FID. Our method consistently improves recall (diversity) but with a slight drop in precision (fidelity). 
As the increase in diversity is dominant, FID, which measures the combined effect of fidelity and diversity, is consistently improved with our method compared to the baselines. Examples of generated samples from our Dia-GAN are also available in the Appendix~\S\ref{supp:images_GAN}.

\paragraph{Experiments on StyleGAN2}
We further evaluate the scalability of our method with StyleGAN2~\cite{stylegan2} on FFHQ 256x256~\cite{stylegan} dataset.
We train the model for 250k steps in total where phase 1 takes 200k steps and the phase 2 takes the remaining steps.
We set the hyperparameter $k=3.0$.
Our method improves the FID of StyleGAN2 from 14.07 to 11.89 and recall of StyleGAN2 from 0.27 to 0.30 as shown in Table~\ref{tab:stylegan2}.
This indicates that our method successfully scales to large state-of-the-art GANs and high-resolution images.

\paragraph{Extension to hinge loss}
We further conduct experiments to show the applicability of our method to other GAN losses.
Here, we focus on a commonly used loss, the hinge loss (HingeGAN)~\cite{hinge1,hinge2}.
Our method is not directly applicable to the hinge loss since the output of the optimal discriminator $D_h(x)$ is not $\frac{\pd(x)}{\pd(x)+p_g(x)}$ anymore. 
Instead, $D_h(x)$ is $1$ if $\pd (x)>p_g(x) $ and $-1$ if $\pd(x)<p_g(x)$. 
One possible workaround is attaching an auxiliary layer to the discriminator and training it with the original GAN loss. 
However, we instead present empirical evidence showing that $D_h(x)$ itself still contains useful information about the degree of learning for the input $x$. 
We consider the variant of our method, Dia-HingeGAN, by calculating the mean and variance of $D_h(x)$ and using the same scoring rule of~\eqref{eqn:discrepancy}.
In Table~\ref{table:hingegan}, we compare the performance of HingeGAN and Dia-HingeGAN with the same configuration of the previous experiment.
Interestingly, our method shows significant improvement in both CIFAR-10 and CelebA.
This implies that despite the optimal form of the discriminator is different, the statistics of its output still provide meaningful information about the underrepresented features.
We leave the theoretical analysis of this variant method as a future work.

\begin{table}[t]
\begin{minipage}{.45\linewidth}
    \caption{StyleGAN2 on FFHQ 256x256.}
    \label{tab:stylegan2}
    \vspace{0.2em}
    \centering
    \begin{tabular}{c|ccc}
        \toprule
        & FID $\downarrow$ & P $\uparrow$   & R $\uparrow$ \\
        \midrule
        StyleGAN2 & 14.07   & \textbf{0.72}  & 0.27 \\
        GOLD    & 15.53 & 0.69  & 0.29 \\
        \textbf{Dia-StyleGAN2}  & \textbf{11.89} & 0.69  & \textbf{0.30} \\
        \bottomrule
    \end{tabular}
\end{minipage} 
\begin{minipage}{.53\linewidth}
    \caption{HingeGAN on CIFAR-10 and CelebA.}
    \label{table:hingegan}
    \vspace{0.2em}
    \centering

    
    
    \begin{tabular}{c|cc|c}
    \toprule \label{tab:hingeGAN}
    \multirow{2}{*}{}     &  \multicolumn{2}{c|}{CIFAR-10} & CelebA\\
    \cmidrule{2-4}
    & FID $\downarrow$ & IS $\uparrow$ & FID $\downarrow$\\
    \midrule
    HingeGAN & 21.99 & 7.67 & 6.66\\
    \textbf{Dia-HingeGAN} & \textbf{18.74} & \textbf{8.02} & \textbf{5.98}\\
    \bottomrule
    \end{tabular}

\end{minipage}
\end{table}
\subsection{Minor feature generation}\label{sec:exp_art}

\paragraph{Controlled experiments}
As our method emphasizes underrepresented samples in GAN training, we evaluate how much our method helps the generation of minor samples. 
To control the level of minority, we design a Colored MNIST dataset with red (major) and green (minor) samples, and MNIST-FNIST dataset with MNIST (major) FMNIST (minor) samples, with the majority rates $\rho\in\{90,95,99\} \%$.
We compare our method with the same set of baseline methods as in Section~\ref{sec:5.1_baselines}.
Additionally, we compare our method with PacGAN~\cite{lin2020pacgan}, the approach to handle the mode collapse problem, and with Inclusive GAN~\cite{InclusiveGAN}, which also improves the data coverage over the minor groups by using a hybrid generative model. 

Table~\ref{table:controlled_recon} shows the results of each method in various majority rates.
Here, we focus on the reconstruction error (RE) score of minor  training samples (green samples for Colored MNIST, and FMINST samples for MNIST-FMNIST dataset).
For the training dataset with the majority rate of 99\%, our method shows a significant improvement in RE score as only our method succeeds in generating minor samples while others fail.
When the majority rate $\rho$ decreases to 95\% and 90\%, vanilla model starts to generate minor samples but in low quality.
For these rates, our method also shows improvement on the quality of generated samples with minor features, resulting in better RE scores.
This result implies the efficacy of our method in improving the quality of generated samples with underrepresented features.
Examples of samples with minor features for each method are available in the Appendix~\S\ref{supp:color_GAN}, and
detailed configuration of the experiment is available in the Appendix~\S\ref{supp:simul}.
\begin{table}[!tb]
\centering
\caption{Reconstruction Error (RE) score of green (minor) training samples in Colored MNIST and FMNIST (minor) samples in a mixture of MNIST and FMNIST on different majority rate $\rho$.}
\label{table:controlled_recon}
\vspace{0.2em}
\begin{tabular}{c|ccc|ccc}
\toprule
Dataset & \multicolumn{3}{c|}{Colored MNIST} & \multicolumn{3}{c}{MNIST-FMNIST}\\
\midrule
Majority rate $\rho$ &{99\%} &{95\%} &{90\%}  &{99\%} &{95\%} &{90\%}\\
\midrule
Vanilla & 0.838  & 0.236  & 0.218 & 0.290 & 0.227 & 0.215\\
GOLD~\cite{GOLD}  & 0.813 & 0.297 & 0.200 & 0.296 & 0.241 & 0.218 \\
Top-k~\cite{sinha2020top}  & 0.831  & 0.210  & 0.223 & 0.281 & 0.232 & 0.221\\
PacGAN~\cite{lin2020pacgan}  & 0.810  & 0.244  & 0.233 & 0.313 & 0.251 & 0.225\\
Inclusive GAN~\cite{InclusiveGAN} & 0.812 & 0.274 & 0.216 & 0.283 & 0.230 & 0.220 \\
\textbf{Dia-GAN}  & \textbf{0.224}&  \textbf{0.204} & \textbf{0.197} & \textbf{0.264} & \textbf{0.219} & \textbf{0.206}\\
\bottomrule
\end{tabular}
\end{table}
\newcolumntype{?}{!{\vrule width 0.8pt}}

\begin{table}[!tb]
\centering
\caption{CelebA minor attribute analysis. Averaged LDRV and averaged discrepancy score of CelebA samples with (W/) or without (W/O) minor attributes. O stands for the occurrence of minor attributes among the generated samples in percentage (\%) and R stands for the Partial Recall.}
\label{table:celeba_attr_count_ldrv}
\vspace{0.2em}
\begin{tabular}{c?cc|cc?cc|cc}
\toprule
\multirow{3}{*}{} &\multicolumn{4}{c?}{Score}&\multicolumn{4}{c}{Method}\\
\cmidrule{2-9}
&\multicolumn{2}{c|}{LDRV }&\multicolumn{2}{c?}{Discrepancy }&\multicolumn{2}{c|}{Vanilla} &\multicolumn{2}{c}{\textbf{Dia-GAN}}\\
\cmidrule{2-9}
& W/ & W/O & W/ & W/O & O $\uparrow$ & R $\uparrow$ & O $\uparrow$ & R $\uparrow$ \\
\midrule
Bald (2.244\%) & \textbf{0.271} & 0.184 & \textbf{2.938} & 2.221& 0.678 & 0.353 & \textbf{0.836} & \textbf{0.393} \\
Double Chin (4.669\%) & \textbf{0.219} & 0.184 & \textbf{2.525} & 2.224 & 0.440 & 0.411 & \textbf{0.522} & \textbf{0.461}\\
Eyeglasses (6.512\%) & \textbf{0.254} & 0.181 & \textbf{2.783} & 2.200 & 3.300 & 0.400 & \textbf{4.053} & \textbf{0.449}\\
Gray Hair (4.195\%) & \textbf{0.211} & 0.185 & \textbf{2.450} & 2.228 & 2.273 & 0.402 & \textbf{2.369} & \textbf{0.436}\\
Mustache (4.155\%) & \textbf{0.242} & 0.183 & \textbf{2.699} & 2.218 & 0.157 & 0.391 & \textbf{0.228} & \textbf{0.433}\\
Pale Skin (4.295\%) & \textbf{0.190} & 0.186 & \textbf{2.240} & 2.238 & 0.346 & 0.380 & \textbf{0.453} & \textbf{0.427}\\
Wearing Hat (4.846\%) & \textbf{0.357} & 0.177 & \textbf{3.651} & 2.164 & 2.307 & 0.380 & \textbf{3.595} & \textbf{0.408}\\
\bottomrule
\end{tabular}
\end{table}

\paragraph{CelebA minor attribute analysis}
For the real-world example, we analyze how our method changes the generation of minor attributes of the CelebA~\cite{CelebA} dataset, using the meta-information available in the dataset.
Specifically, we focus on how much our method improves the occurrence rate of minor attributes, as they usually appear in a much lower rate than its actual ratio.
We train a binary classifier for each attribute to have train and test accuracy above 95\%.
We also evaluate the Partial Recall of the minor attributes, since minor samples suffer low-recall problem as explained in Section~\ref{subsec:ldrv}.

\begin{figure}[t]
    \centering
    \includegraphics[width=\linewidth]{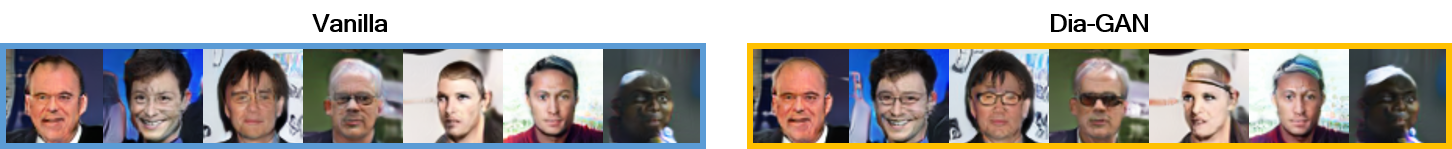}
    \caption{Generated samples of vanilla SNGAN and Dia-SNGAN trained on CelebA.}
    \label{fig:celeba_gen_images}
\end{figure}

As shown in Table~\ref{table:celeba_attr_count_ldrv}, our method improves both the occurrence (O) and Partial Recall (R) rates of various minor attributes.
Moreover, as explained in Section~\ref{subsec:ldrv}, minor samples do have higher LDRV~\eqref{eqn:LDRMLDRV} and the discrepancy score~\eqref{eqn:discrepancy}.
This indicates that our method indeed captures the underrepresented minor features and successfully promotes the generation of such features during training of GANs. 
Figure~\ref{fig:celeba_gen_images} shows examples of generated samples with minor feature appeared by our Dia-GAN.
Note that as we use the majority of the training time for Phase 1 (80\% of total steps), the generator partially converges after Phase 1 and thus the same latent vector $z$ turns out to give a similar image with details changed after Phase 2 (e.g., wearing sunglasses, or a hat).

The ability of our method in capturing semantic features and improving the generation of minor samples also applies to high-resolution datasets. To demonstrate this, we conduct similar experiments for the high-resolution FFHQ dataset and present the results in the Appendix~\S\ref{sec:FFHQ}.

\section{Discussion}
\label{sec:discussion}
We proposed two new metrics, LDRV and LDRM, that can detect underrepresented samples and devised a simple approach to emphasize detected underrepresented samples.
Our method successfully improves overall quality of the generated samples in terms of FID and IS, and promotes generation of minor samples.
However, we still find the trade-off relationship between the precision and recall of generated samples (Table~\ref{table:cifar10-celeba-results}).
We leave the investigation of other approaches to use the knowledge of detected underrepresented samples for further improvement of GAN training as a future work.

\paragraph{Societal impact}
We propose a discrepancy score that can detect underrepresented minor samples in training of GANs.
On the good side, this results in enhanced generation of minor samples in GANs.
The ability of our score could be further expanded and used for utilizing skewed datasets to train models representing more balanced datasets, by adding a hyperparameter that can tune the level of emphasis for underrepresented samples.
On the other hand, an abuser might instead be able to remove such minor subgroup samples and deteriorate the bias in sample generation.

\section*{Acknowledgement}
This research was supported by the National Research Foundation of Korea under Grant 2017R1E1A1A01076340 and 2021R1C1C11008539, and by the Ministry of Science and ICT, Korea, under the IITP (Institute for Information and Communications Technology Panning and Evaluation) grant (No.2020-0-00626).

\newpage
\bibliography{main}
\bibliographystyle{plain}

 \newpage
 \appendix
 \renewcommand{\thesection}{\Alph{section}}

\numberwithin{equation}{section}
\numberwithin{thm}{section}
\setcounter{table}{0}
\renewcommand{\thetable}{A\arabic{table}}
\setcounter{figure}{0}
\renewcommand{\thefigure}{A\arabic{figure}}

\setcounter{section}{0}

\section{Instability of LDR estimate}\label{supp:LDR}

\begin{figure}[!htb]
    \centering
    \subfloat[\label{fig:LDR_instability}]{\includegraphics[width=0.35\linewidth]{{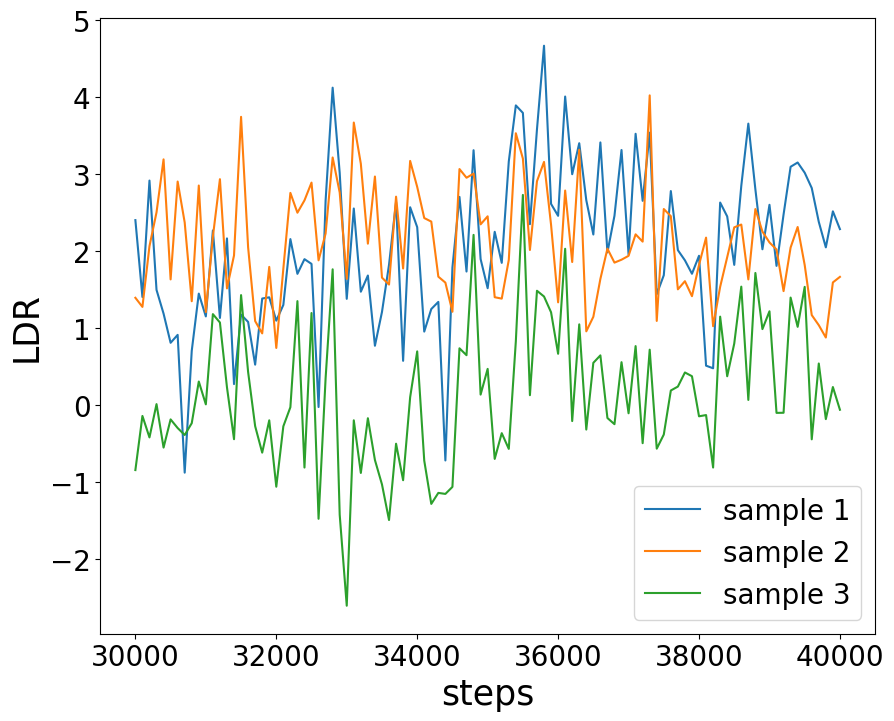}}} 
	\subfloat[\label{fig:LDR_plot}]{\includegraphics[width=0.65\linewidth]{{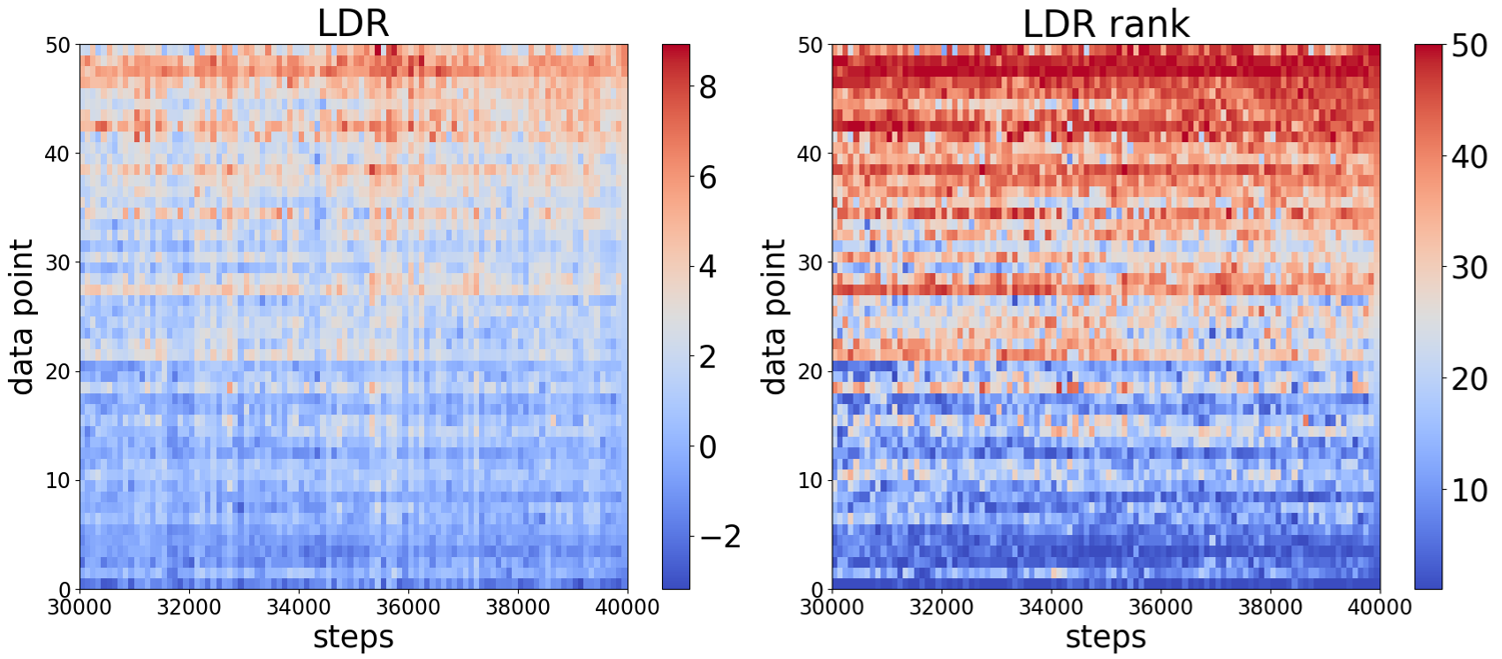}}}
	\caption{(a) LDR values of three randomly chosen samples, (b) LDR values (left) and rankings (right) of 50 samples during training of CIFAR-10 on SNGAN~\cite{sngan}. The values are recorded every 100 steps from 30000 to 40000 steps (total 100 times). LDR values are unstable during training, so it is hard to diagnose GAN training from the LDR of a particular training step. The level of fluctuations varies much over samples. 
	}\label{fig:LDR_inst}
\end{figure}
The Log-Density-Ratio estimate (LDR) is defined by
\beq\label{eqn:ALDR}
\LDR(x):=\log\frac{D(x)}{1-D(x)}.
\eeq
When $D(x)=D^{*}(x)$, the $\LDR(x)$ is equal to the log density ratio $\log (\pd(x)/p_g(x))$. 
When $\LDR(x)> 0$, the data point $x$ is underrepresented in the model, i.e., $\pd(x)>p_g(x)$, while when $\LDR(x)< 0$, the data is overrepresented, i.e., $\pd(x)<p_g(x)$.
Thus, we can leverage the value of $\LDR(x)$ of each instance $x$ to give feedback to improve the generator if the estimation is valid.
Some prior works have used this tendency to evaluate the quality of {fake samples} and designed sample reweighting scheme to guide the generator to focus on underestimated samples \cite{GOLD} or rejection sampling to post-process generated samples to approximately correct errors in the model distribution \cite{DRS}.

The effectiveness of the above schemes highly depends on the accuracy of the LDR estimate. However, we observe that $\LDR(x)$ is unstable during the training even after large steps, as shown in Fig.~\ref{fig:LDR_inst}. 
Therefore, to have a better estimate on LDR, we propose to use statistics (mean and variance) of LDR estimates over multiple steps (epochs) of the training.
Different from \cite{DRS, GOLD}, we focus on the discrepancy of GANs at training data instances rather than at generated samples. 
This leads us to fully explore the underrepresented regions of the data manifold during the training, which can then be emphasized to improve the performance of GANs.

\section{Analysis of variance of LDR estimate}\label{supp:analysis}

Consider the discriminator trained with a data set $\{({x}_i,y_i)_{i=1}^n\}$ to minimize the cross-entropy loss
\beq
-\sum_{i=1}^n \left(y_i \log D({x}_i)+(1-y_i)\log (1-  D({x}_i)) \right)
\eeq
where $y_i=1$ for a real sample and $y_i=0$ for a fake sample. Assuming that $\mathbf{\phi}_i=F(x_i)\in \mathbb{R}^d$ denotes the feature vector of $x_i$ extracted by the discriminator and that the discriminator is defined by a sigmoid applied to $\theta^T\phi_i$ for some $d$-dimensional parameter $\theta$ just like the logistic regression, the discriminator output can be considered as the probability that the input $x_i$ is a real sample, i.e.,
\beq\label{eqn:appD_likelihood}
D({x}_i;\theta)=\frac{1}{1+e^{-{\theta}^T \phi_i }}=p(y_i=1|\phi_i,\theta).
\eeq
We now turn to a Bayesian treatment of logistic regression and find the Gaussian approximation for the posterior distribution of $\theta$ given the data set, in a similar way as in Section 4.5 of~\cite{bishop2006pattern}.
Assume that 
\beq\label{eqn:theta_prior}
p(\theta)=\mathcal{N}(\theta|0, s_0 I)
\eeq
where $s_0$ is a fixed hyperparameter. The posterior distribution over $\theta$ is given by
\beq
p(\theta|(\phi_i,y_i)_{i=1}^n)\propto p(\theta)p(y_1^n|\phi_1^n, \theta).
\eeq
Taking the log of both sides, and substituting for the prior distribution~\eqref{eqn:theta_prior}, and the likelihood function using~\eqref{eqn:appD_likelihood}, we obtain
\beq
\log p(\theta|(\phi_i,y_i)_{i=1}^n)=\sum_{i=1}^n \left(y_i \log D({x}_i;\theta)+(1-y_i)\log (1-  D({x}_i;\theta)) \right) -\frac{\|\theta\|^2}{2s_0} +\text{const.}
\eeq
for $ D({x}_i;\theta)$ in~\eqref{eqn:appD_likelihood}.
To obtain a Gaussian approximation to the posterior distribution, we first find $\theta_{\sf MAP}$ that maximizes $\log p(\theta|(\phi_i,y_i)_{i=1}^n)$, i.e., $\frac{d}{d\theta} \log p(\theta_d|(\phi_i,y_i)_{i=1}^n)\Big|_{\theta=\theta_{\sf MAP}} =0$, which defines the mean of the Gaussian. The covariance is then given by the inverse of the matrix of second derivatives of the negative log likelihood, which takes the form
\beq
S_n^{-1}= -\bigtriangledown_{\theta} \bigtriangledown_{\theta} \log p(\theta|(\phi_i,y_i)_{i=1}^n)=\sum_{i=1}^n D({x}_i;\theta)(1-D({x}_i;\theta)) \phi_i \phi_i^T +\frac{1}{s_0}I.
\eeq
The Gaussian approximation of the posterior distribution of $\theta$ thus takes the form of
\beq\label{p_post}
p(\theta|(\phi_i,y_i)_{i=1}^n)\approx \mathcal{N}(\theta|\theta_{\sf MAP}, S_n).
\eeq
We next relate the variance of LDR estimate for each data sample with the covariance matrix $S_n$. 
First, we can find that
\beq
\begin{split}
\var\left(\log\frac{D({x}_i;\theta)}{1-D({x}_i;\theta)}\right)=\var(\log D)+\var(\log (1-D))-2\text{cov}(\log D, \log(1-D)).
\end{split}
\eeq
By approximating $D({x}_i;\theta)$ by the Taylor expansion at $\theta=\theta_{\sf MAP}$, we get
\beq
\begin{split}
\log D({x}_i;\theta)&\approx \log D({x}_i; \theta_{\sf MAP})+(1-D({x}_i;\theta_{\sf MAP}))\phi_i^T(\theta-\theta_{\sf MAP}),\\
\log (1-D(x_i;\theta))&\approx \log (1-D({x}_i; \theta_{\sf MAP}))-D( {x}_i;\theta_{\sf MAP})\phi_i^T(\theta-\theta_{\sf MAP}),
\end{split}
\eeq
and thus the variances are
\beq
\begin{split}
\var(\log D({x}_i;\theta))&\approx (1-D({x}_i;\theta_{\sf MAP}))^2 \phi_i^TS_n \phi_i,\\
\var(\log (1-D({x}_i;\theta)))&\approx D({x}_i;\theta_{\sf MAP})^2 \phi_i^TS_n\phi_i,
\end{split}
\eeq
for the covariance matrix $S_n$ of~\eqref{p_post}.
Using the similar Taylor expansion, we can approximate
\beq
\begin{split}
\text{cov}(\log D, \log(1-D))&=\E[\log D\cdot \log(1-D)]-\E[\log D]\E[\log(1-D)]\\
&\approx-D({x}_i;\theta_{\sf MAP})(1-D({x}_i;\theta_{\sf MAP}))\phi_i^TS_n\phi_i.
\end{split}
\eeq
By combining the above results,
\beq
\begin{split}\label{eqn:LDRV_ind}
\var\left(\log\frac{D({x}_i;\theta)}{1-D({x}_i;\theta)}\right)&\approx (D^2+(1-D)^2+2D(1-D))\phi_i^TS_n\phi_i\\
&=\phi_i^TS_n\phi_i.
\end{split}
\eeq
Finally, by plugging in $S_n$, the variance of LDR estimate for each sample ${x}_i$ with feature vector $\phi_i$ can be written as
\beq
\var\left(\log\frac{D({x}_i;\theta)}{1-D({x}_i;\theta)}\right)\approx \phi_i^T \left(\sum_{i=1}^n D({x}_i;\theta)(1-D({x}_i;\theta)) \phi_i \phi_i^T +\frac{1}{s_0}I\right)^{-1} \phi_i.
\eeq

\section{Images with lowest/highest discrepancy score for CIFAR-10 \& CelebA}\label{supp:images_DS}
\begin{figure*}[!htb]
\centering
\begin{tabular}{cc}
	\subfloat[Images with lowest disc. score \label{fig:celeba_small}]{\includegraphics[width=0.32\linewidth]{{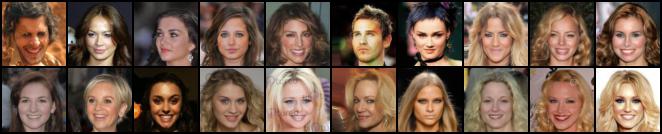}}} &\\

    \subfloat[Images with highest disc. score \label{fig:celeba_large}]{\includegraphics[width=0.32\linewidth]{{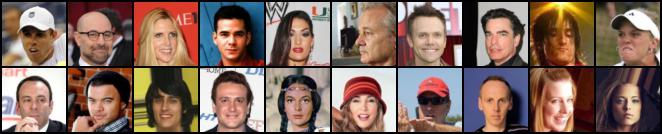}}}&\\
    
    \subfloat[Generated samples\label{fig:celeba_gen}]{\includegraphics[width=0.32\linewidth]{{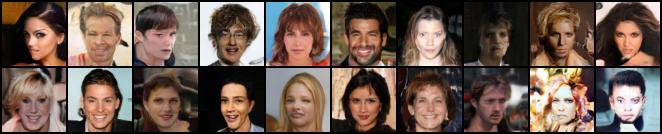}}}&  
    \multirow{-12.8}{*}{    \subfloat[Historgram of pixel count over intensity level    \label{fig:celeba_histogram}]{\includegraphics[width=0.44\linewidth,height=0.32\linewidth]{{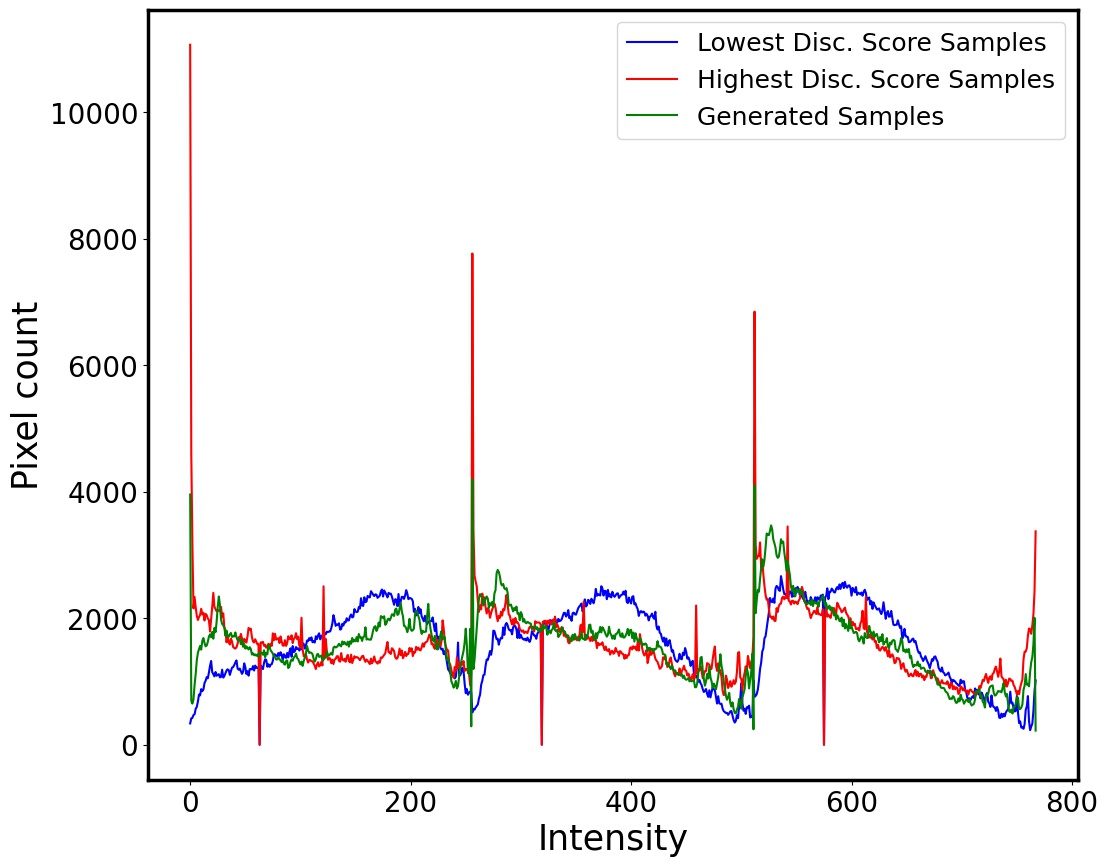}}}}\\
    \end{tabular}
    \caption{Examples of CelebA training images with (a) lowest and (b) highest discrepancy scores, and (c) generated samples after Phase 1 (without weighted sampling). Generated samples resemble training images with lowest discrepancy score. (d) A smoothed histogram of the intensities for 100 samples per group.
    The intensity levels of RGB channels are concatenated, resulting in a total of $768=256\times3$ levels. 
    Images with the lowest scores (blue) and generated samples (green) have a similar distribution, while images with the highest scores (red) show a high discrepancy. 
    }
    \label{fig:celeba_samples}
\end{figure*}
In this section, we show the characteristics of training images having lowest/highest discrepancy scores. 
As of Fig. \ref{fig:cifar10_samples} (CIFAR-10) in the main paper, we present the images with lowest (Fig.~\ref{fig:celeba_small})/highest (Fig. \ref{fig:celeba_large}) discrepancy scores among CelebA training images, and compare them with generated samples (Fig.~\ref{fig:celeba_gen}), after Phase 1 of our algorithm (before sample-weighting starts).
Comparing the pixel intensity histogram (Fig.~\ref{fig:celeba_histogram}) reveals more clearly the difference in sample properties.
Images with low discrepancy scores exhibit similar intensity distribution with generated samples, while images with high scores appear to show an extremely different tendency.
These results show that our discrepancy score successfully distinguishes underrepresented instances.

\begin{figure}[!ht]
    \centering
    \subfloat[Samples with high LDRM ($k$=0)]{\includegraphics[width=0.3\linewidth]{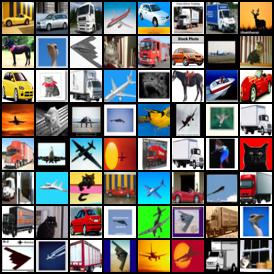}} \quad
    \subfloat[Samples with high LDRV ($k\to\infty$) ]{\includegraphics[width=0.3\linewidth]{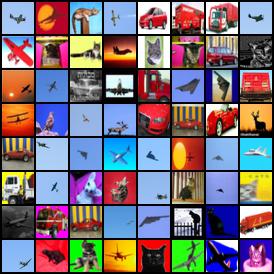}}\\
    \subfloat[Samples with high disc. score ($k$=0.3)]{\includegraphics[width=0.3\linewidth]{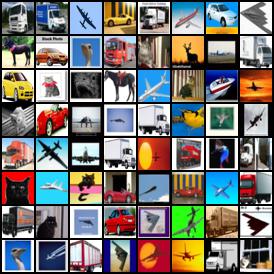}} \quad
    \subfloat[Samples with high disc. score ($k$=0.5)]{\includegraphics[width=0.3\linewidth]{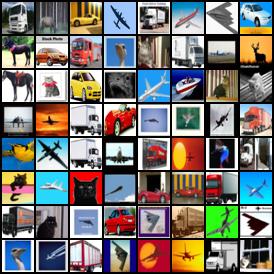}}\quad
    \subfloat[Samples with high disc. score ($k$=1.0)]{\includegraphics[width=0.3\linewidth]{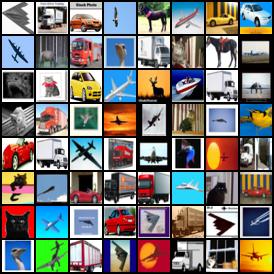}}\\
    \subfloat[Samples with high disc. score ($k$=3.0)]{\includegraphics[width=0.3\linewidth]{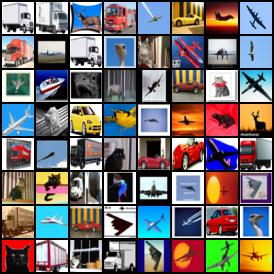}}\quad
    \subfloat[Samples with high disc. score ($k$=5.0)]{\includegraphics[width=0.3\linewidth]{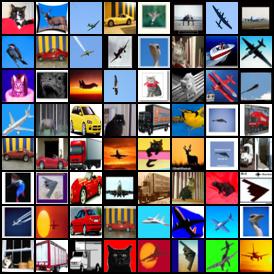}}\quad
    \subfloat[Samples with high disc. score ($k$=7.0)]{\includegraphics[width=0.3\linewidth]{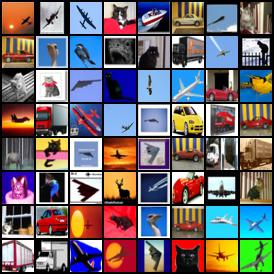}}
    \caption{CIFAR-10 samples with highest discrepancy scores on various $k$}
    \label{fig:CIFAR-10_high_samples}
\end{figure}

\begin{figure}[!ht]
    \centering
    \subfloat[Samples with low LDRM ($k$=0)]{\includegraphics[width=0.3\linewidth]{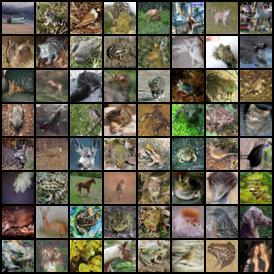}} \quad
    \subfloat[Samples with low LDRV ($k\to\infty$) ]{\includegraphics[width=0.3\linewidth]{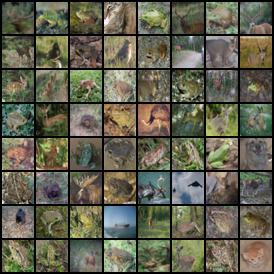}}\\
    \subfloat[Samples with low disc. score ($k$=0.3)]{\includegraphics[width=0.3\linewidth]{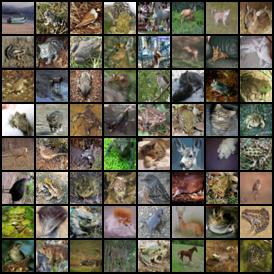}} \quad
    \subfloat[Samples with low disc. score ($k$=0.5)]{\includegraphics[width=0.3\linewidth]{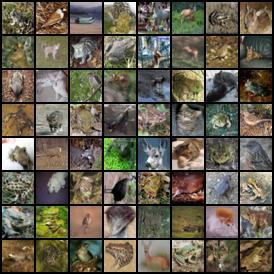}}\quad
    \subfloat[Samples with low disc. score ($k$=1.0)]{\includegraphics[width=0.3\linewidth]{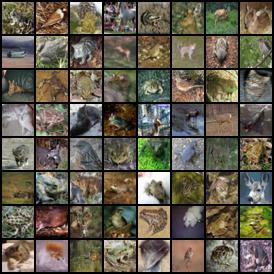}}\\
    \subfloat[Samples with low disc. score ($k$=3.0)]{\includegraphics[width=0.3\linewidth]{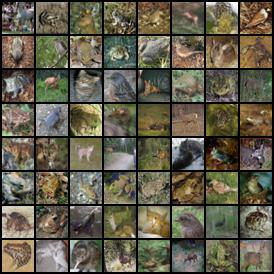}}\quad
    \subfloat[Samples with low disc. score ($k$=5.0)]{\includegraphics[width=0.3\linewidth]{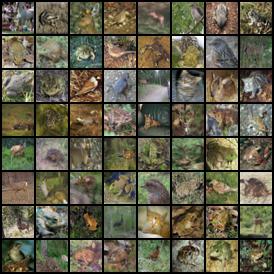}}\quad
    \subfloat[Samples with low disc. score ($k$=7.0)]{\includegraphics[width=0.3\linewidth]{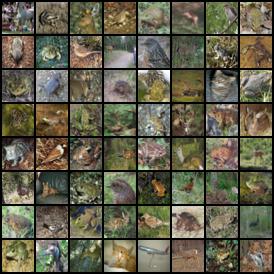}}
    \caption{CIFAR-10 samples with lowest discrepancy scores on various $k$}
    \label{fig:CIFAR-10_low_samples}
\end{figure}

\begin{figure}[!ht]
    \centering
    \subfloat[Samples with high LDRM ($k$=0)]{\includegraphics[width=0.3\linewidth]{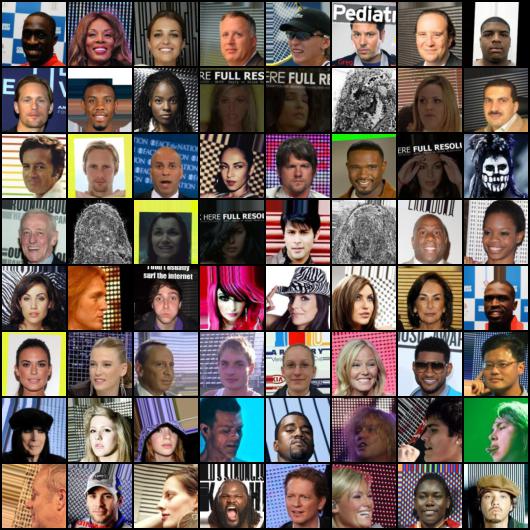}} \quad
    \subfloat[Samples with high LDRV ($k\to\infty$) ]{\includegraphics[width=0.3\linewidth]{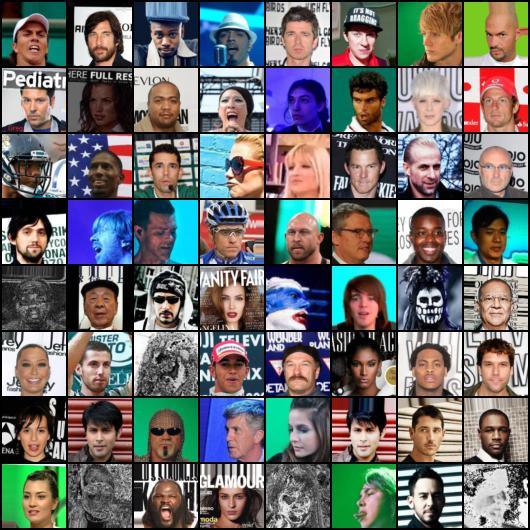}}\\
    \subfloat[Samples with high disc. score ($k$=0.3)]{\includegraphics[width=0.3\linewidth]{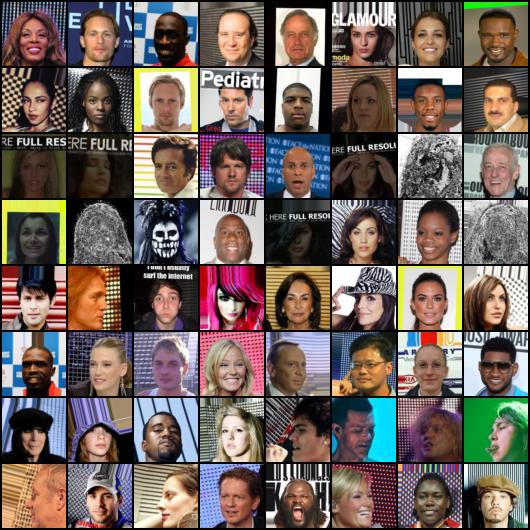}} \quad
    \subfloat[Samples with high disc. score ($k$=0.5)]{\includegraphics[width=0.3\linewidth]{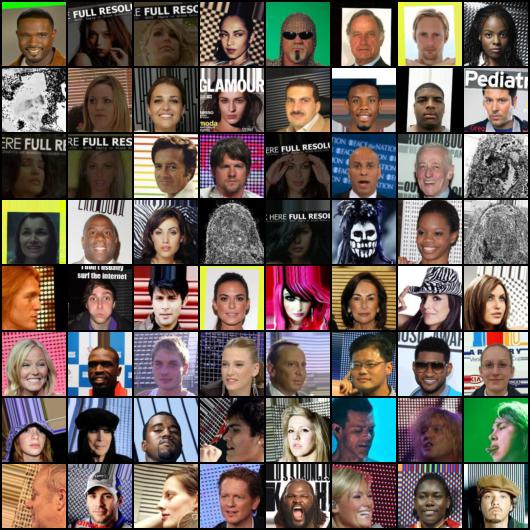}}\quad
    \subfloat[Samples with high disc. score ($k$=1.0)]{\includegraphics[width=0.3\linewidth]{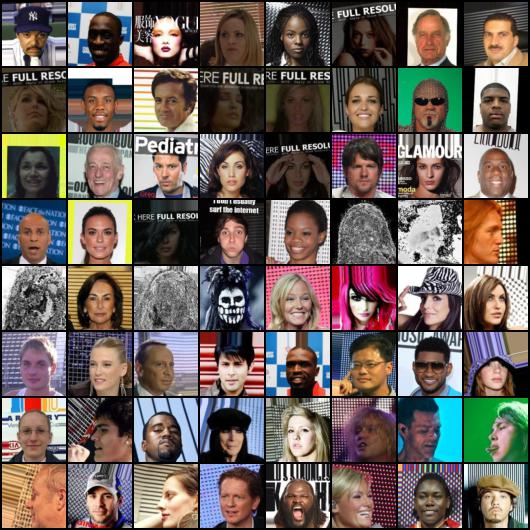}}\\
    \subfloat[Samples with high disc. score ($k$=3.0)]{\includegraphics[width=0.3\linewidth]{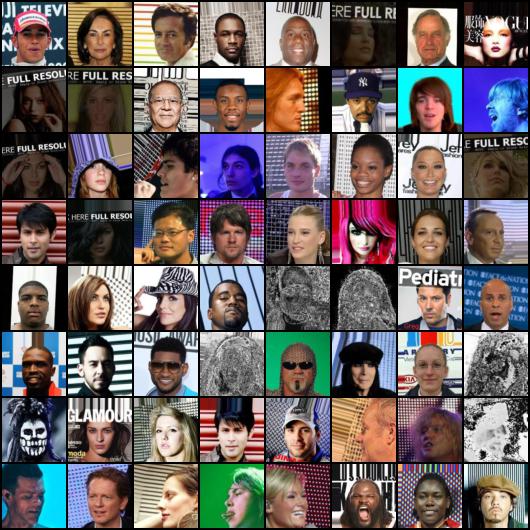}}\quad
    \subfloat[Samples with high disc. score ($k$=5.0)]{\includegraphics[width=0.3\linewidth]{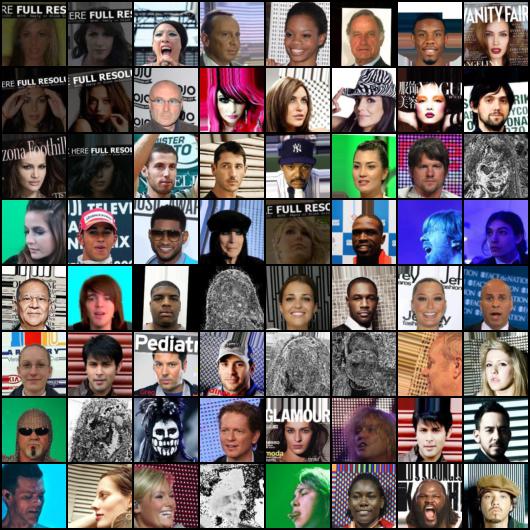}}\quad
    \subfloat[Samples with high disc. score ($k$=7.0)]{\includegraphics[width=0.3\linewidth]{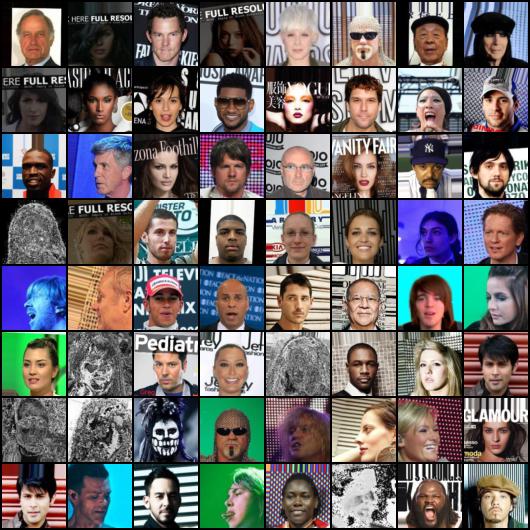}}
    \caption{CelebA samples with highest discrepancy scores on various $k$}
    \label{fig:CelebA_high_samples}
\end{figure}

\begin{figure}[!ht]
    \centering
    \subfloat[Samples with low LDRM ($k$=0)]{\includegraphics[width=0.3\linewidth]{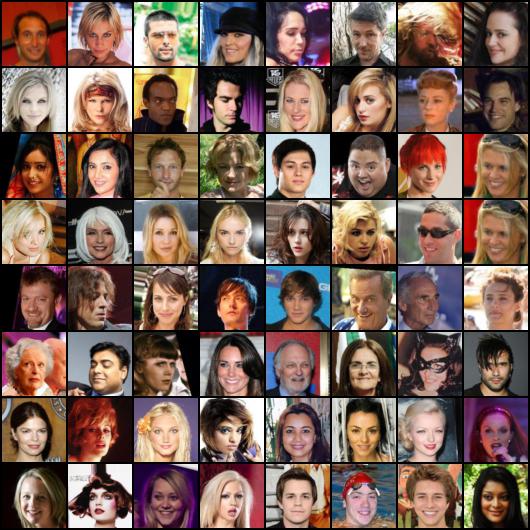}} \quad
    \subfloat[Samples with low LDRV ($k\to\infty$)]{\includegraphics[width=0.3\linewidth]{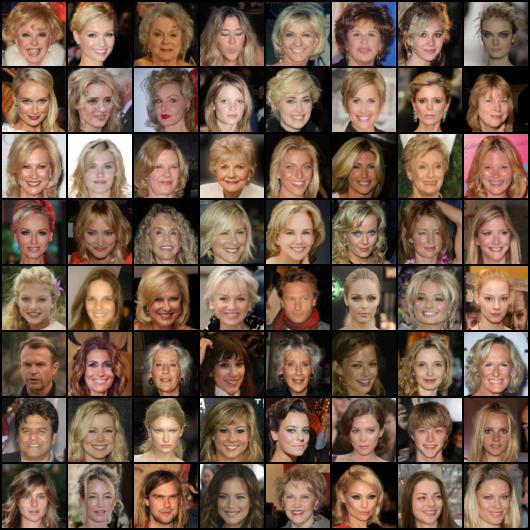}}\\
    \subfloat[Samples with low disc. score ($k$=0.3)]{\includegraphics[width=0.3\linewidth]{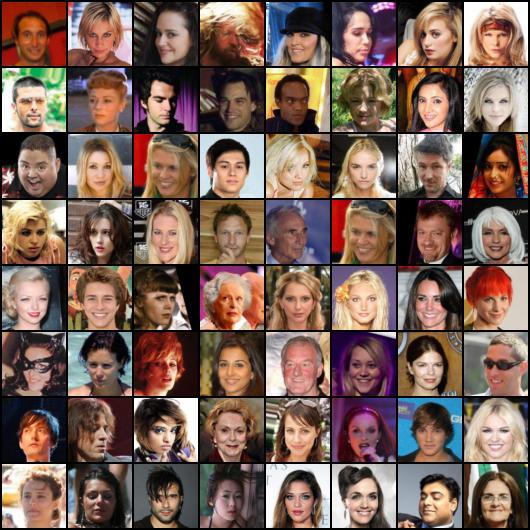}}\quad
    \subfloat[Samples with low disc. score ($k$=0.5)]{\includegraphics[width=0.3\linewidth]{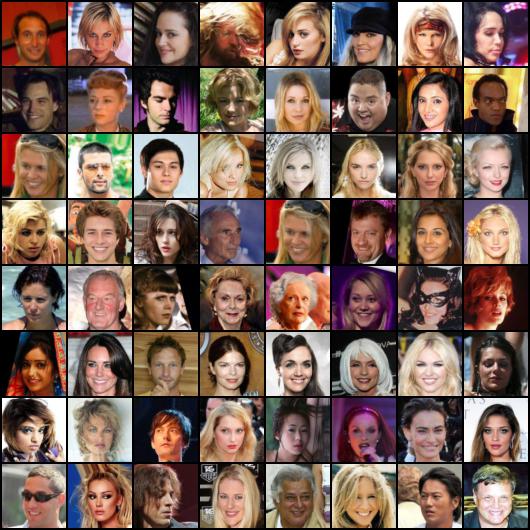}}\quad
    \subfloat[Samples with low disc. score ($k$=1.0)]{\includegraphics[width=0.3\linewidth]{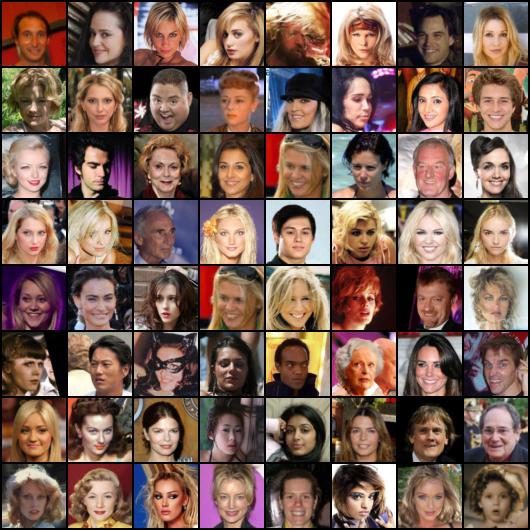}}\\
    \subfloat[Samples with low disc. score ($k$=3.0)]{\includegraphics[width=0.3\linewidth]{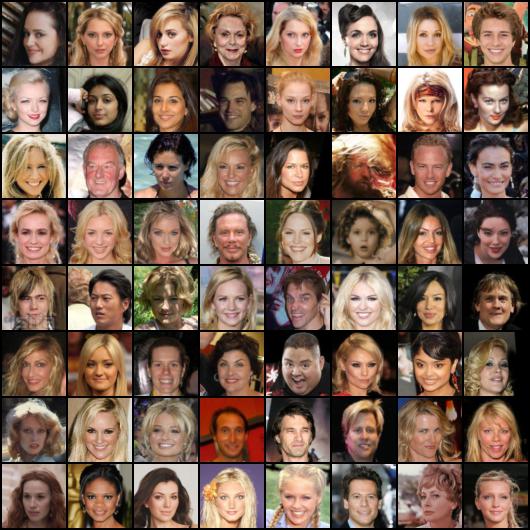}}\quad
    \subfloat[Samples with low disc. score ($k$=5.0)]{\includegraphics[width=0.3\linewidth]{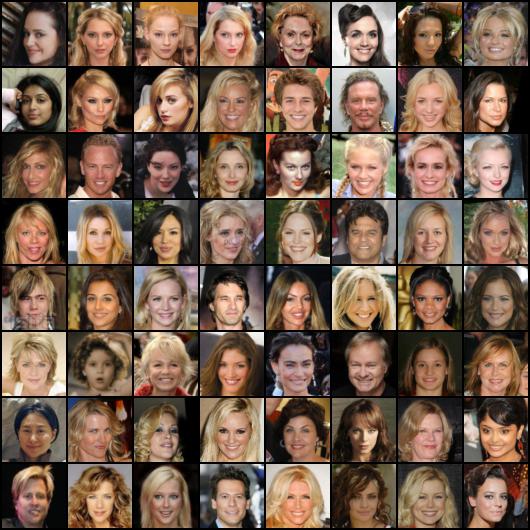}}\quad
    \subfloat[Samples with low disc. score ($k$=7.0)]{\includegraphics[width=0.3\linewidth]{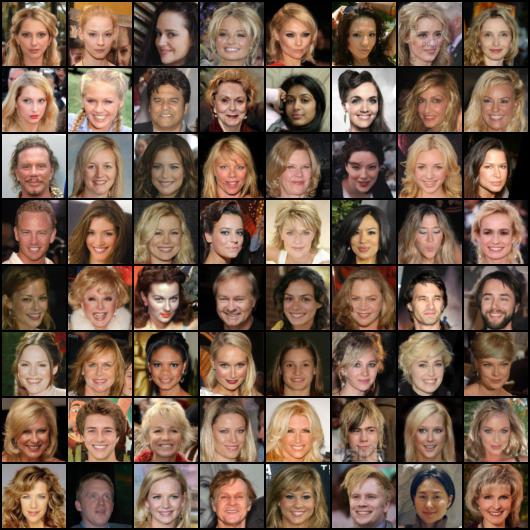}}
    \caption{CelebA samples with lowest discrepancy scores on various $k$}
    \label{fig:CelebA_low_samples}
\end{figure}

We also present the samples with lowest/highest discrepancy scores with various $k$ values (the hyperparameter for discrepancy score~\eqref{eqn:discrepancy}) for CIFAR-10 (Figure \ref{fig:CIFAR-10_high_samples}, \ref{fig:CIFAR-10_low_samples}) \& CelebA (\ref{fig:CelebA_high_samples}, \ref{fig:CelebA_low_samples}). 
Samples with high discrepancy scores have properties that are distinct from the samples with low discrepancy scores (e.g. vividness or unusual backgrounds for CIFAR-10 and minor features such as diverse hair colors or accessories including glasses or hats for CelebA).
Since generated samples resemble the images with low discrepancy scores, emphasizing high-scoring images can boost the diversity in sample generation. 


\clearpage

\section{Algorithm}\label{supp:alg}
Detailed algorithm description for Self-Diagnosing GAN is introduced in Algorithm \ref{alg:resample}.
\begin{algorithm}[!htb]
		\caption{Self-Diagnosing GAN(Dia-GAN)}
	\label{alg:resample}
	\begin{algorithmic}
		\STATE {\bfseries Input:} Dataset $\mathcal{D}$, Model $\mathcal{M}=\{D,G,D_{\sf aux}\}$, Batch size $B$, Numbers of steps for phase 1 and 2 ($t_1$ and $t_2$), step to start recording LDR $t_s$, Number of samples to be generated $N$
		\STATE {\bfseries Output:} Set of generated samples $\{g_1, g_2, \cdots, g_N\}$ 
		\STATE \textbf{Phase 1 - Train and Diagnose} 
		\STATE Initialize $\theta_D^0, \theta_G^0$
		\FOR{$t \leftarrow 1$ {\bfseries to} $t_1$}
		\STATE $\cD_{B}^{t} \leftarrow \{x_{i}: x_{i} \sim \text{Unif}(\cD)\}$
		\STATE $\cZ^t \leftarrow\{G(z_j): z_j\sim p_z(z)\}$
		\STATE $\theta_D^t \leftarrow \theta_D^{t-1}+\eta_D \nabla_{\theta_D} V_{\sf D}(D,G;\cD_{B}^t, \cZ^t )$ for $V_D$ in~\eqref{ns_disc} (NS loss) or~\eqref{hinge_disc} (hinge loss)
		\STATE $\theta_G^t \leftarrow \theta_G^{t-1}-\eta_G\nabla_{\theta_G} V_{\sf G}(D, G;\cZ^t )$ for $V_G$ in~\eqref{ns_gen} (NS loss) or~\eqref{hinge_gen} (hinge loss).
		
		\IF  {$t \ge t_s$}
		\STATE Save $\LDR(x_i)_t$ for $i \in \{1,2, \cdots, |\mathcal{D}|\}$
		\ENDIF
		
		\ENDFOR
		\STATE Compute discrepancy score $s(x_i;\{t_s, t_s + 1, \cdots , t_1\})$  for $i \in \{1,2, \cdots, |\mathcal{D}|\}$. (Eq. \eqref{eqn:discrepancy})
		\STATE Compute sampling frequency $P_s(i)\propto  \texttt{max\_clip}(\texttt{min\_clip}(s(x_i;T)))$, for $i \in \{1,2, \cdots, |\mathcal{D}|\}$.
		\STATE \textbf{Phase 2 - Score-Based Weighted Sampling} 
		\STATE Initialize $\theta_{D_{\sf aux}}^{t_1} \leftarrow \theta_D^{t_1}$
		\FOR{$t\leftarrow t_1 + 1$ {\bfseries to} $t_1 +t_2$}
		\STATE $\cD_{B}^{t} \leftarrow \{x_{i}: x_{i} \sim P_s(i)\}$
		\STATE ${\cD_B^t}' \leftarrow \{x_{i}: x_{i} \sim \text{Unif}(\cD)\}$
		\STATE $\cZ^t \leftarrow\{G(z_j): z_j\sim p_z(z)\}$
		\STATE $\theta_D^t \leftarrow \theta_D^{t-1}+\eta_D \nabla_{\theta_D} V_{\sf D}(D,G;\cD_{B}^t, \cZ^t)$
		\STATE $\theta_G^t \leftarrow \theta_G^{t-1}-\eta_G \nabla_{\theta_G} V_{\sf G}(D, G;\cZ^t )$
		\STATE $\theta_{D_{\sf aux}}^t \leftarrow \theta_{D_{\sf aux}}^{t-1}+\eta_{D_{\sf aux}}\nabla_{\theta_{D_{\sf aux}}} V_{\sf D}(D_{\sf aux},G;{\cD_{B}^t}', \cZ^t)$
		\ENDFOR
		\STATE \textbf{Phase 3 - DRS} 
		\STATE $\{g_1, g_2, \cdots, g_N\} \leftarrow {\sf DRS}(G; D_{\sf aux}, N)$
	\end{algorithmic}
\end{algorithm}

\paragraph{Algorithm complexity}
Compared to the original GAN training, the overhead in time and space from our method is not dominant.
For CIFAR-10 dataset, 5 hours 38 minutes were required to train 50k steps of Dia-SNGAN (our method), while 4 hours 51 minutes were needed for the original SNGAN in RTX 3090 GPU.
Similarly, for CelebA dataset, 19 hours 53 minutes were required to train 75k steps of Dia-SNGAN (our method), while 17 hours 7 minutes were needed for the original SNGAN with the same GPU.
Diagnosing samples in Phase 1 requires additional space for saving LDR values. 
Phase 2 needs additional auxiliary discriminator training to perform DRS in Phase 3. 
However, this does not require much overhead since Phase 2 is shorter than Phase 1 and we initialize auxiliary discriminator using the original discriminator trained in Phase 1.

\section{Variants of the original GAN loss and description of evaluation metrics}\label{supp:eval}
\subsection{Non-saturating GAN loss}
We consider a practical training method of GANs, using alternative SGD, to solve $\min_G\max_D V(D,G)$ for $V(D,G)=\E_{x\sim \pd}[ \log D(x)]+\E_{z\sim p_z} [\log(1-D(G(z)))]$. The mini-batches of $B$ samples for the training dataset and fake samples are defined as $\cD_B=\{x^{(j)}: x^{(j)}=x_i \text{ where } i\sim P_s(i) \text{ for } j=1,\dots, B\}$ and $\cZ=\{G(z^{(j)}): z^{(j)}\sim p_z(z) \text{ for } j=1,\dots, B\}$, respectively.
Then, the alternative training of GAN updates the discriminator parameter $\theta_D$ and the generator parameter $\theta_G$  by backpropagating the gradient of GAN loss calculated on these mini-batches.

For training, we use the non-saturating variant of the generator loss,
\begin{align}
	&V_{\sf D}(D,G;\cD_{B}, \cZ )= \frac{1}{|\cD_{B}|}\sum _{\cD_{B}} \log D(x^{(j)}) +\frac{1}{|\cZ|}\sum_{\cZ} \log (1-D(G(z^{(j)}))), \label{ns_disc}\\
	&V_{\sf G}(D, G;\cZ )  =-\frac{1}{|\cZ|}\sum_{\cZ} \log D(G(z^{(j)})).   \label{ns_gen}
\end{align}

\subsection{Hinge GAN loss}

Several types of GANs achieve enhanced performance when the hinge loss~\cite{hinge1,hinge2} is applied instead of the normal non-saturating loss. 
In Section~\ref{sec:exp_real}, we demonstrate the applicability of our score-based weighted sampling to GANs with hinge loss.
The hinge loss is defined as 
\begin{align}
	&V_{\sf D}(D,G;\cD_{B}, \cZ ) = \frac{1}{|\cD_{B}|} \sum _{\cD_{B}} \min (0, -1+D(x^{(j)})) +\frac{1}{|\cZ|} \sum_{\cZ} \min (0, -1-D(G(z^{(j)}))), \label{hinge_disc}\\
	&V_{\sf G}(D, G;\cZ ) =- \frac{1}{|\cZ|} \sum_{\cZ} D(G(z^{(j)})).\label{hinge_gen}
\end{align}

\subsection{Description of evaluation metrics}
To evaluate the effect of our method on learned model distribution, we use various evaluation metrics including (1) Fr\'echet Inception Distance (FID)~\cite{heusel2017gans}, (2) Inception Score (IS)~\cite{salimans2016improved}, and (3) Precision and Recall (P\&R)~\cite{PR}. 
\begin{itemize}
\item FID measures the distance between the model distribution and the data distribution, approximated by two multidimensional Gaussian distributions in the feature space of InceptionV3~\cite{Szegedy_2016_CVPR} classifier, so it measures the overall fitness of the model distribution to the data distribution, in terms of both the quality (fidelity) and the diversity. 
\item IS measures the quality of generated samples, in the sense that whether the generated samples can be classified by InceptionV3 classifier into each of the dataset classes. 
\item Precision is described as the portion of generated samples that fall within the data manifold, which measures the fidelity of generated samples, while recall measures the portion of data instances falling within the manifold of generated samples, which measures the diversity. We follow the definitions of precision and recall by \cite{PR}, which are described as follows.

Let the set of feature vectors of real and generated samples be $\Phi_r$, $\Phi_g$, respectively. Also, let the binary function $f(\phi, \Phi)$ be
\begin{equation}
f(\phi, \Phi) = \begin{cases}
1, & \text{$\exists \phi' \in \Phi$ s.t. $\left\|\phi-\phi'\right\|_{2} \leq\left\|\phi'-\mathrm{NN}_{k}\left(\phi', \Phi\right)\right\|_{2}$ } \\
0, &\text{otherwise}
\end{cases}
\end{equation}
where $\mathrm{NN}_{k}(\phi', \Phi)$ denotes the $k-$th nearest feature vector of $\phi'$ in set $\Phi$. Then, precision and recall is defined as:
\begin{align}
    \mathtt{precision} &= \frac{1}{\left|{\Phi}_{g}\right|} \sum_{{\phi}_{g} \in {\Phi}_{g}} f\left({\phi}_{g}, {\Phi}_{r}\right) \\
    \mathtt{recall} &= \frac{1}{\left|{\Phi}_{r}\right|} \sum_{{\phi}_{r} \in {\Phi}_{r}} f\left({\phi}_{r}, {\Phi}_{g}\right) 
\end{align}

Partial Recall is proposed to measure the recall rate for a subset of dataset. 
It is defined as the portion of data instances in the subset that fall within the manifold of generated samples. 
Let us denote a subset of data and the feature space of that subset by $S$ and $\Phi_S$, respectively. Note that $\Phi_S \subset \Phi_r$.  Then, the partial recall of the subset $S$ is defined as
\begin{equation}
    \mathtt{partial\_recall}(S) = \frac{1}{\left|{\Phi}_{S}\right|} \sum_{{\phi}_{S} \in {\Phi}_{S}} f\left({\phi}_{S}, {\Phi}_{g}\right) 
\end{equation}

\end{itemize}
In addition to these global evaluation metrics, to evaluate whether a subset of dataset is well represented in the model distribution, we consider (4) Reconstruction Error (RE)~\cite{anomaly_drae,anomaly_rdae}.  
\begin{itemize}
\item Reconstruction Error (RE) score is calculated by training a convolutional autoencoder (CAE) with generated samples and then calculating the Euclidean distance between each training data and its reconstruction.
RE can assess whether $p_g(x)$ covers $\pd(x)$ since CAE is known to have high RE for out-of-distribution samples~\cite{anomaly_drae,anomaly_rdae}.
RE score for a subset of data is defined as the average RE score of each data instance within the subset.
Let us denote the subset of data for which we want to measure RE score by $S$ and the set of generated samples by $F$. The autoencoder output function, trained with samples in $F$, is denoted by $\theta_F(\cdot)$. Then, the RE score of a subset $S$ is defined as 
\begin{equation}
    \texttt{RE}(S) = \frac{1}{|S|} \sum_{s \in S} {\texttt{dist} (\theta_F(s),s)} 
\end{equation}
where $\texttt{dist}(x,y)$ measures the distance between $x$, $y$ and is defined as the Euclidean distance averaged for all pixels. 

\end{itemize}

For data statistics to calculate FID, we use provided results for CIFAR-10\footnote{\url{http://bioinf.jku.at/research/ttur/}} and calculate statistics for CelebA with all the training samples. We compare the statistics of 50,000 generated samples with these data statistics. We use 50,000 generated samples to evaluate IS, and 10,000 data samples and 10,000 generated samples to evaluate P\&R. Also, we use the feature layer of the InceptionV3 classifier instead of the feature layer of VGG16 as in \cite{PR}.

\section{Details of simulation setups}\label{supp:simul}
\subsection{Controlled dataset - single-mode Gaussian}
We generate 2-D single-mode Gaussian dataset with mean $\mathbf{0}$ and various covariance $\sigma \mathbf{I}$.
We use $\sigma\in\{3, 2.5, 2\}$.
Minority level (Fig. \ref{fig:recall_plot}) 1, 2, 3 stands for $\sigma = $ 3, 2.5, 2, respectively.  
As $\sigma$ decreases, the samples concentrate more near the mean of the Gaussian, and this aligns with the situation that minority rate decreases in the Colored MNIST dataset or the MNIST-FMNIST mixture dataset.
The size of the dataset is 10,000.
We use a GAN architecture based on the multi-layer perceptron (MLP) with details described in Table \ref{table:mlp}. 
We use the batch size of 1024 and Adam optimizer with hyperparameters $\alpha=0.001, \beta_1=0.5, \beta_2=0.9$.
We train the model for 200 epochs and record LDR for every sample at the end of each epoch while training.
We define the major group as the samples within distance two from the origin, and the minor group as the samples outside of distance seven from the origin.
To compute Partial Recall, we use data itself as the feature.
Fig \ref{fig:recall_plot} and Table \ref{tab:minor_ldrv} are the experimental results averaged from 10 random seeds.

\subsection{Controlled dataset - 25 Gaussian dataset}
We construct the mixture of 25 Gaussians dataset, each centered at $(c_x,c_y)$ for $c_x,c_y \in \{-2,-1,0,1,2\}/1.414$. Each $(x,y) \in \cD$ is sampled from
\beq
(x,y) = \frac{(d_x, d_y) + (z_x, z_y)}{2.828}
\eeq
where $d_x, d_y \in \{-4,-2,0,2,4\}$ and $z_x, z_y \sim N(0,0.05^2)$. The size of the dataset is 10,000, where 400 samples are sampled from each mixture mode. 
We use the same GAN architecture as the one used in the single-mode Gaussian experiment (Table \ref{table:mlp}).
We use the batch size of 128 and Adam optimizer with hyperparameters $\alpha=0.0002, \beta_1=0.5, \beta_2=0.999$. 
We train the model for 300 epochs and record LDR for every sample at the end of each epoch while training.
\begin{table}[!ht]
    \caption{GAN architecture for Single-mode Gaussian and 25 Gaussian dataset}
    \label{table:mlp}
    \centering
    \begin{tabular}{cc}
        \small
        \begin{tabular}{c|c|c}
\toprule
\multicolumn{3}{c}{\textbf{Generator}} \\
\midrule
Layer & Output size & Activation  \\
\midrule
Input $z$ & 2 & \\
FC & 512 & ReLU\\
FC & 512 & ReLU \\
FC & 512 & ReLU \\
FC & 2 &  \\
\bottomrule
\end{tabular}
 &
        \small
        \begin{tabular}{c|c|c}
\toprule
\multicolumn{3}{c}{\textbf{Discriminator}} \\
\midrule
Layer & Output size & Activation  \\
\midrule
Input $x$ & 2 & \\
FC & 512 & ReLU\\
FC & 512 & ReLU \\
FC & 512 & ReLU \\
FC & 1 & Sigmoid \\
\bottomrule
\end{tabular}
    \end{tabular}

\end{table}

 \subsection{Controlled dataset - Colored MNIST \& MNIST-FMNIST}
We generate Colored MNIST by randomly picking 60,000 samples and separating them into two groups corresponding to each color (red and green) at a given majority rate $\rho$.
For the mixture of MNIST and FMNIST dataset, we randomly pick 60,000 samples in total from MNIST and FMNIST dataset with a given majority rate.
For both datasets, minority level (Fig. \ref{fig:recall_plot}) 1, 2, 3 stands for the majority rate $\rho = $ 90\%, 95\%, 99\%, respectively.  
We use DCGAN~\cite{dcgan} with the detailed architecture described in Table \ref{table:dcgan_gen} and \ref{table:dcgan_disc}. 
We use the batch size of 64 and Adam optimizer~\cite{adam} with hyperparameters $\alpha=0.0001, \beta_1=0.5, \beta_2=0.9$. 
We additionally use the linear learning rate scheduler that decays until the end of the training.
All models are trained for 20k steps.
For PacGAN~\cite{lin2020pacgan}, we use a packing degree of two.
For Inclusive GAN~\cite{InclusiveGAN}, we use Inception feature~\cite{salimans2016improved} for the feature space.
For GOLD~\cite{GOLD} and our method, the phase 1 takes 15k steps, and the phase 2 takes 5k steps.
For our method, we record LDR every 100 steps and use the last 50 records for calculating the discrepancy score.
We use $k=3$ for Colored MNIST and $k=6$ for MNIST-FMNIST.

To evaluate Partial Recall in Fig. \ref{fig:recall_plot}, we train convolutional classifier (Table \ref{table:convnet}) with 60,000 samples (30,000 major samples and 30,000 minor samples) with 20 classes (Major 10 classes + Minor 10 classes) and use output of flatten layer of this network for the feature space. 
The convolutional classifier is trained for 50 epochs with Adam optimizer~\cite{adam} with hyperparameters $\alpha=0.001, \beta_1=0.9, \beta_2=0.999$ and learning rate scheduler with learning rate decay 0.1 in 42 epoch.
To evaluate reconstruction error (RE) in Table \ref{table:controlled_recon}, we use convolutional autoencoder with the detailed architecture described in Table \ref{table:encoder}, \ref{table:decoder}.
\texttt{nc} in each table states the number of channel.
\texttt{nc} for Colored MNIST is three and \texttt{nc} for MNIST-FMNIST is one.
Fig \ref{fig:recall_plot} and Table \ref{tab:minor_ldrv}, \ref{table:controlled_recon} are the experimental results averaged from three random seeds.

\begin{table}[h]
    \caption{Generator architecture for Colored MNIST \& MNIST-FMNIST}
    \label{table:dcgan_gen}
    \centering
    \small
    \begin{tabular}{c|c|c|c|c|c|c}
    \toprule
    \multicolumn{7}{c}{\textbf{Generator}} \\
    \midrule
    Layer & Output size & Kernel size & Stride & Padding & Batch Norm & Activation  \\
    \midrule
    Input $z$ & 100 & & & & &\\
    FC & 384 & - & - & - & X & \\
    Reshape & 1$\times$1$\times$384 & - & - & - & - & - \\
    Deconv & 4$\times$4$\times$192 & 4$\times$4 & 1 & 0 & O & ReLU \\
    Deconv & 8$\times$8$\times$96 & 4$\times$4 & 2 & 1 & O & ReLU \\
    Deconv & 16$\times$16$\times$48 & 4$\times$4 & 2 & 1 & O & ReLU \\
    Deconv & 32$\times$32$\times$\texttt{nc} & 4$\times$4 & 2 & 1 & X & Tanh \\
    \bottomrule
    \end{tabular}
\end{table}

\begin{table}[ht]
    \caption{Discriminator architecture for Colored MNIST \& MNIST-FMNIST}
    \label{table:dcgan_disc}
    \centering
    \small
    \begin{tabular}{c|c|c|c|c|c|c|c}
    \toprule
    \multicolumn{8}{c}{\textbf{Discriminator}} \\
    \midrule
    Layer & Output size & Kernel size & Stride & Padding & Batch Norm & Dropout & Activation \\
    \midrule
    Input $x$ & 32$\times$32$\times \texttt{nc}$ &     &   &   &   &  \\
    Conv & 16$\times$16$\times$16 & 3$\times$3 & 2 & 1 & X & 0.5 & LeakyReLU(0.2) \\
    Conv & 16$\times$16$\times$32 & 3$\times$3 & 1 & 1 & O & 0.5 & LeakyReLU(0.2) \\
    Conv & 8$\times$8$\times$64 & 3$\times$3 & 2 & 1 & O & 0.5 & LeakyReLU(0.2) \\
    Conv & 8$\times$8$\times$128 & 3$\times$3 & 1 & 1 & O & 0.5 & LeakyReLU(0.2) \\
    Conv & 4$\times$4$\times$256 & 3$\times$3 & 2 & 1 & O & 0.5 & LeakyReLU(0.2) \\
    Conv & 4$\times$4$\times$512 & 3$\times$3 & 1 & 1 & O & 0.5 & LeakyReLU(0.2) \\
    Flatten & - & - & - & - & - & - & - \\
    FC & 1  & - & - & - & X & & Sigmoid\\
    \bottomrule
    \end{tabular}

\end{table}

\begin{table}[t]
    \caption{Classifier architecture for measuring Partial Recall of Colored MNIST \& MNIST-FMNIST}
    \label{table:convnet}
    \centering
    \small
    \begin{tabular}{c|c|c|c|c|c|c}
    \toprule
    \multicolumn{7}{c}{\textbf{Classifier}} \\
    \midrule
    Layer & Output size & Kernel size & Stride & Padding & Batch Norm & Activation  \\
    \midrule
    Input $x$ & 32$\times$32$\times$\texttt{nc} &     &   &   &   \\
    Conv & 32$\times$32$\times$16 & 7$\times$7 & 1 & 3 & O & ReLU \\
    Conv & 32$\times$32$\times$32 & 7$\times$7 & 1 & 3 & O & ReLU \\
    Conv & 32$\times$32$\times$64 & 7$\times$7 & 1 & 3 & O & ReLU \\
    Conv & 32$\times$32$\times$128 & 7$\times$7 & 1 & 3 & O & ReLU \\
    AvgPool & 1$\times$1$\times$128 & - & - & - & - & - \\
    Flatten & - & - & - & - & - & - \\
    FC & 20  & - & - & - & X & Softmax\\
    \bottomrule
    \end{tabular}

\end{table}

\begin{table}[t]
    \caption{Encoder architecture for measuring Reconstruction Error (RE) score}
    \label{table:encoder}
    \centering
    \small
    \begin{tabular}{c|c|c|c|c|c|c}
    \toprule
    \multicolumn{7}{c}{\textbf{Encoder}} \\
    \midrule
    Layer & Output size & Kernel size & Stride & Padding & Batch Norm & Activation  \\
    \midrule
    Input $x$ & 32$\times$32$\times$\texttt{nc} &     &   &   &   \\
    Conv & 16$\times$16$\times$64 & 3$\times$3 & 2 & 1 & O & ReLU \\
    Conv & 8$\times$8$\times$128 & 3$\times$3 & 2 & 1 & O & ReLU \\
    Conv & 4$\times$4$\times$256 & 3$\times$3 & 2 & 1 & O & ReLU \\
    Flatten & - & - & - & - & - & - \\
    FC & 256  & - & - & - & X & Tanh\\
    \bottomrule
    \end{tabular}

\end{table}

\begin{table}[!ht]
    \caption{Decoder architecture for measuring Reconstruction Error (RE) score}
    \label{table:decoder}
    \centering
    \small
    \begin{tabular}{c|c|c|c|c|c|c|c}
    \toprule
    \multicolumn{8}{c}{\textbf{Decoder}} \\
    \midrule
    Layer & Output size & Kernel size & Stride & Padding & Output padding & Batch Norm & Activation \\
    \midrule
    Input $z$ & 256 & & & & & &\\
    FC & (4$\times$4$\times$256) & - & - & - & - & O & ReLU \\
    Reshape & 4$\times$4$\times$256 & - & - & - & - & - & - \\
    Deconv & 8$\times$8$\times$128 & 3$\times$3 & 2 & 1 & 1 & O & ReLU \\
    Deconv & 16$\times$16$\times$64 & 3$\times$3 & 2 & 1 & 1 & O & ReLU \\
    Deconv & 32$\times$32$\times$\texttt{nc} & 3$\times$3 & 2 & 1 & 1 & X & Tanh \\
    \bottomrule
    \end{tabular}

\end{table}

\subsection{Real dataset - CIFAR-10 and CelebA}
We evaluate our method with two types of GANs: SNGAN~\cite{sngan} and SSGAN~\cite{ssgan} \footnote{When we train SSGAN with the Top-k method, we only consider top-$k$ samples for the GAN tasks, while we use full (not top-$k$) samples for the self-supervised tasks.}. 
Following~\cite{sngan}, we use the residual network architecture proposed in ResNet~\cite{resnet} for all GAN variants.
Our experimental code is based on the GAN research library Mimicry~\cite{lee2020mimicry}.
We use batch size of 64 and Adam optimizer~\cite{adam} with hyperparameters $\alpha=0.0002, \beta_1=0, \beta_2=0.9$. The learning rate is set to decay linearly with the training steps. 
Table \ref{table:cifar10-celeba-results} and \ref{table:hingegan} are the experimental results averaged from three random seeds.

\subsection{Real dataset - FFHQ}
We test the scalability of our method on the large-scale model.
Specifically, we train StyleGAN2~\cite{stylegan2} on FFHQ 256x256~\cite{stylegan} dataset.
We follow most of the techniques used in the original StyleGAN2~\cite{stylegan2}.
We use leaky ReLU activation with $\alpha=0.2$, bilinear filtering~\cite{zhang2019invariant} in all up/downsampling layers, minibatch standard deviation layer at the end of
the discriminator~\cite{karras2018progressive}, exponential moving average of generator weights~\cite{karras2018progressive} and style mixing regularization~\cite{stylegan}.
For the discriminator, we use $r_1=0.1$ for the weight of $R_1$ regularizer~\cite{lars2018gan} and apply the lazy regularization~\cite{stylegan2} every 16 steps.
For the generator, we apply path length regularization~\cite{stylegan2} with weight of 2 and batch size reducing factor of 2 and also apply the lazy regularization every 4 steps.
We use the batch size of 16 and Adam optimizer~\cite{adam} with the hyperparameters $\alpha=0.0016, \beta_1=0, \beta_2=0.991$. 
In total, we train for 250k where the phase 1 takes 200k steps and phase 2 takes the remaining 50k steps.
We record the LDR values every 100 steps for the last 5k steps of phase 1 (195k $\sim$ 200k).
For the discrepancy score, we use $k=3.0$.
Table \ref{tab:stylegan2} shows the experimental results averaged from two random seeds.

\subsection{Details on CelebA minor attribute analysis}
To analyze the CelebA minor attribute, we use the meta-information provided by CelebA~\cite{CelebA}. We use a pre-trained VGG16 network to train attribute classifiers for each attribute. Except for the last three fully connected layers, we fix the parameters of the pre-trained VGG16 network and change the output size of last layer to two. We train only the last three layers (classifier layers) of the VGG16 network for 10 epochs with batch size 128 and SGD optimizer with a learning rate of 0.001 and momentum of 0.9. We select the minor attributes with accuracy above 95\% for train and test datasets. We count the occurrence of minor attributes using the trained classifier.
Table \ref{table:celeba_attr_count_ldrv} is the experimental results averaged from three random seeds.

\subsection{Hyperparameter search for discrepancy score}
The hyperparmeter $k$ for discrepancy score~\eqref{eqn:discrepancy} is chosen from $k=0.3, 0.5, 1.0, 3.0, 5.0, 7.0$ to achieve the best FID score among the candidates for each dataset at SNGAN, and the value of $k$ is fixed across the GAN variants. See Table \ref{Atable:hyperparameter} for details. As we can see in Table \ref{Atable:hyperparameter}, an appropriate choice of $k$ can be different depending on the dataset. These are results averaged from three random trials.

\begin{table*}[!htb]
    \caption{FID for Dia-GAN with various $k$} 
    \label{Atable:hyperparameter}
    \centering
    \small
    \begin{tabular}{c|c|c|c|c|c|c}
    \toprule
    $k$    & 0.3    & 0.5 & 1.0    & 3.0 & 5.0 & 7.0\\
    \midrule
    FID for CIFAR-10 & \textbf{19.23} & 19.47 & 20.58 & 23.45 & 24.44 & 20.43\\
    \midrule
    FID for CelebA & 7.27 & 6.91 & 6.52 & 6.73 & \textbf{6.37} & 6.41\\
    \bottomrule
    \end{tabular}

\end{table*}

\subsection{Hyperparameter choice for training steps}
We choose the training steps for Phase 1 of our algorithm as 80\% of total steps to make sure that the discriminator is trained enough. However, experiments with the different training step choices shown in Table \ref{Atable:hyperparameter_steps} imply our method's robustness on the parameter choice. 
\begin{table}[h]
    \caption{FID for Dia-GAN with different phase 1 steps (\% of total steps).}
    \label{Atable:hyperparameter_steps}
    \centering
    \small
    \begin{tabular}{c|c|c|c|c|c}
    \toprule
    & \textbf{Baseline} & \textbf{20\%} & \textbf{40\%} & \textbf{60\%} & \textbf{80\%} \\
    \midrule
    FID for CIFAR-10 & 26.90\scriptsize{$\pm$0.90} & 17.56\scriptsize{$\pm$1.03} & \textbf{16.72\scriptsize{$\pm$0.74}} & 18.65\scriptsize{$\pm$0.94} & 19.66\scriptsize{$\pm$0.42} \\
    \midrule
    FID for CelebA & 7.12\scriptsize{$\pm$0.27} & \textbf{6.69\scriptsize{$\pm$0.33}} & 6.90\scriptsize{$\pm$0.66} & 6.86\scriptsize{$\pm$0.77} & 6.70\scriptsize{$\pm$0.69}\\
    \bottomrule
    \end{tabular}

\end{table}

When we take the longer total training steps as 100k steps for SNGAN on CIFAR-10 and CelebA, we find similar trends as we use 50k steps for total training steps. See Table \ref{table:longer_training} for details. The overall FID gets better when the model is trained longer, but our method still gives an improvement in term of FID, Inception score and recall. In addition, we want to point out that our method can offer an efficient way of training, as our method requires much fewer steps to achieve FID better that the best FID of the Vanilla GAN.
\newcolumntype{?}{!{\vrule width 0.8pt}}

\begin{table}[!tb]
\centering
\caption{FID for SNGAN and Dia-SNGAN with different total training steps.}
\label{table:longer_training}
\vspace{0.2em}
\small
\begin{tabular}{c?c|c?c|c|c}
\toprule
\multirow{2}{*}{} & \multicolumn{2}{c?}{CIFAR-10} & \multicolumn{3}{c}{CelebA} \\
\cmidrule{2-6}
& FID $\downarrow$ & IS $\uparrow$ & FID $\downarrow$ & P $\uparrow$ & R $\uparrow$ \\
\midrule
SNGAN (50k/ 75k) & 26.90\scriptsize{$\pm$0.90} & 7.36\scriptsize{$\pm$0.08} & 7.12\scriptsize{$\pm$0.27} & \textbf{0.68\scriptsize{$\pm$0.00}} & 0.44\scriptsize{$\pm$0.01} \\
Dia-SNGAN (50k/ 75k) & \textbf{19.66\scriptsize{$\pm$0.42}} & \textbf{7.95\scriptsize{$\pm$0.09}} & \textbf{6.70\scriptsize{$\pm$0.69}} & 0.64\scriptsize{$\pm$0.02} & \textbf{0.48\scriptsize{$\pm$0.02}} \\
\midrule
SNGAN (100k) & 22.43\scriptsize{$\pm$0.92} & 7.59\scriptsize{$\pm$0.06} & 6.83\scriptsize{$\pm$0.46} & \textbf{0.68\scriptsize{$\pm$0.00}} & 0.45\scriptsize{$\pm$0.02} \\
Dia-SNGAN (100k) & \textbf{16.49\scriptsize{$\pm$1.05}} & \textbf{8.10\scriptsize{$\pm$0.14}} & \textbf{6.57\scriptsize{$\pm$0.70}} & 0.63\scriptsize{$\pm$0.01} & \textbf{0.49\scriptsize{$\pm$0.01}} \\
\bottomrule
\end{tabular}
\end{table}

\subsection{Necessity of combining LDRM and LDRV}\label{supp:LDRMV}
Our discrepancy score is the weighted sum of two metrics, balancing the effects of two terms.
To check the effects of combining two metrics, we train SNGAN on CIFAR-10 and CelebA using only LDRM or LDRV metric.
We use clipped LDRM or clipped LDRV value as we applied to the discrepancy score.
As shown in Table \ref{Atable:LDRM_LDRV_weighted_sampling}, average FID of using only one metric cannot achieves average FID of using discrepancy score.
This implies the importance of incorporating LDRV over LDRM and the effect of proper balancing of both metrics.
\begin{table}[!h]
    \caption{FID for GAN using weighted sampling with LDRM and LDRV.}
    \label{Atable:LDRM_LDRV_weighted_sampling}
    \centering
    \small
    \begin{tabular}{c|c|c|c|c}
    \toprule
    & \textbf{Baseline} & \textbf{LDRM} & \textbf{LDRV} & \textbf{Dia-GAN(Ours)} \\
    \midrule
    FID for CIFAR-10 & 26.90\scriptsize{$\pm$0.90} & 19.80\scriptsize{$\pm$0.47} & 20.08\scriptsize{$\pm$0.67} & \textbf{19.66\scriptsize{$\pm$0.42}} \\
    \midrule
    FID for CelebA & 7.12\scriptsize{$\pm$0.27} & 7.46\scriptsize{$\pm$0.57} & 7.08\scriptsize{$\pm$0.75} & \textbf{6.70\scriptsize{$\pm$0.69}} \\
    \bottomrule
    \end{tabular}
\end{table}

\subsection{Details on Discriminator Rejection Sampling (DRS)}
In this subsection, we introduce the practical scheme of DRS by briefly referring to original DRS paper and explain our hyperparameter uses for DRS algorithm. 
Discriminator Rejection Sampling \cite{DRS} accepts the fake sample $x$ with probability $\pd(x)/Mp_g(x)$  where $M = \max_{x}\left( \pd(x)/p_g(x)\right)$. If we let $\LDR_M = \log M$, then acceptance probability for $x$, denoted by $p_{\mathrm {accept}}(x)$, would be
\beq
    p_{\mathrm {accept}}(x) = e^{\LDR(x)- \LDR_M}.
\eeq
To deal with low acceptance probabilities and numerical stability issue, Azadi et al. \cite{DRS} instead proposed to compute $F(x)$ which satisfies
\beq
    p_{\mathrm {accept}}(x) = \frac{1}{1+ e^{-F(x)}}.
\eeq
Equivalently,
\beq
    F(x) = \LDR(x) - \LDR_M - \log(1- e^{\LDR(x) - \LDR_M}).
\eeq
Practically, in DRS algorithm we compute
\beq
    \hat{F}(x) = \LDR(x) - \LDR_M - \log(1- e^{\LDR(x) - \LDR_M - \epsilon}) -\gamma,
\eeq
where $\epsilon$  is a constant for preventing overflow and $\gamma$ is a hyperparameter for controlling the acceptance probability.
For applying DRS with auxiliary discriminator in our algorithm, we used $\epsilon = 10^{-6}$ and let $\gamma $ be 80\% percentile of $\hat{F}(x)$. $\LDR_M$ is initially estimated with $256 \times 50 = 12800$ samples by finding the maximum LDR value among those samples. $\LDR_M$ is updated during sampling phase whenever a bigger one is found.

\section{Effect of our method in sample generation for CIFAR-10 \& CelebA}\label{supp:images_GAN}
\subsection{Visualized effect of weighted sampling}
\begin{figure}[!ht]
    \centering
    \subfloat[Generated samples with original sampling method]{\includegraphics[width=0.45\linewidth]{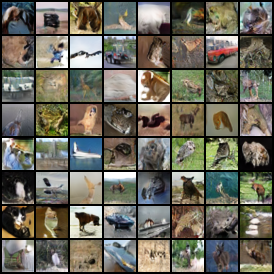}} \quad
    \subfloat[Generated samples with weighted sampling method (ours)]{\includegraphics[width=0.45\linewidth]{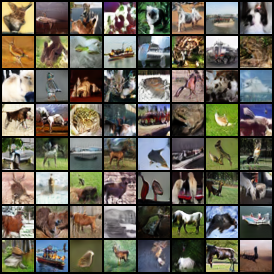}}
    \caption{Example of generated samples with (a) original sampling and with (b) weighted sampling (CIFAR-10)}
    \label{fig:gen_cifar10}
\end{figure}

\begin{figure}[!ht]
    \centering
    \subfloat[Generated samples with original sampling method]{\includegraphics[width=0.45\linewidth]{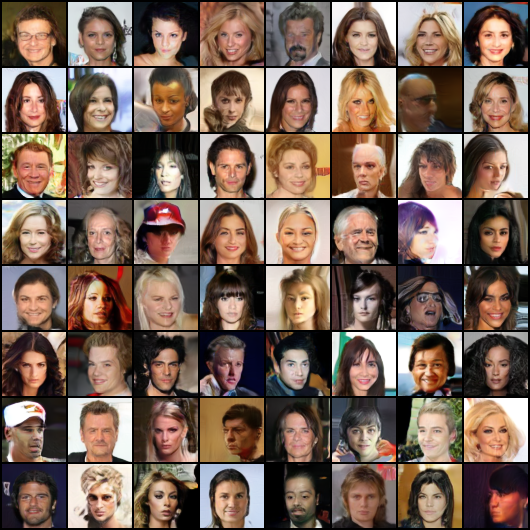}} \quad
    \subfloat[Generated samples with weighted sampling method (ours)]{\includegraphics[width=0.45\linewidth]{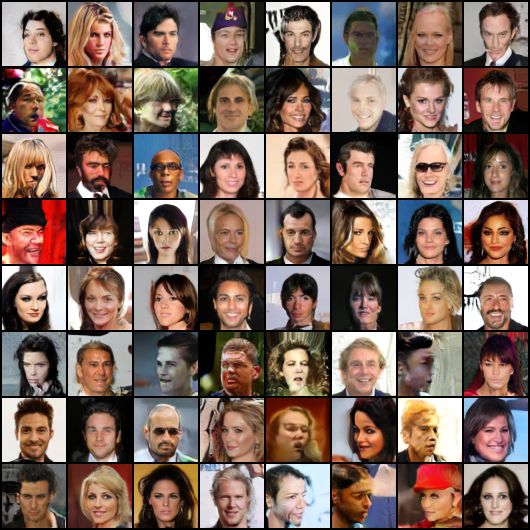}}
    \caption{Example of generated samples with (a) original sampling and with (b) weighted sampling (CelebA)}
    \label{fig:gen_celeba}
\end{figure}

\begin{figure*}[htb!]
\centering
\begin{tabular}{cc}       
    \subfloat{\includegraphics[scale=0.5]{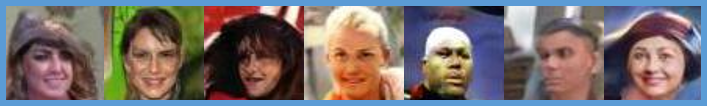}} & \subfloat{\includegraphics[scale=0.5]{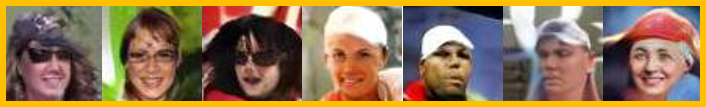}} \\
    \subfloat{\includegraphics[scale=0.5]{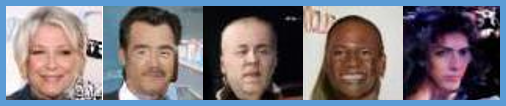}} & \subfloat{\includegraphics[scale=0.5]{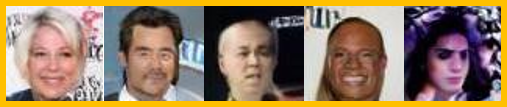}} \\ 
    (a) Generated samples with original sampling & (b) Generated samples with weighted sampling
\end{tabular}
\caption{Comparison of generated samples with (a) original sampling and with (b) weighted sampling for CelebA dataset. Examples of generated images with minor features such as accessories including glasses or hats appeared by our method (1st row), or with more unique backgrounds (e.g. with some letters in the background) (2nd row).}
\label{fig:weighted_sampling}
\end{figure*}

In Fig. \ref{fig:gen_cifar10}  (CIFAR-10) and \ref{fig:gen_celeba} (CelebA), we compare the generated samples with and without our sampling method, which emphasizes underrepresented samples having high discrepancy scores. 
We also visualize the effect of our weighted sampling by showing some examples of generated samples $G(z)$ with the same $z$ between original GAN and our method in  Fig. \ref{fig:weighted_sampling}. 
In  Fig. \ref{fig:weighted_sampling}, we show some examples of CelebA images with minor features such as accessories including glasses or hats appeared by our weighted sampling, and also images having unique backgrounds (e.g. with some letters in the background) with our method.
These minor features are often underrepresented in sample generation of original GANs, while our weighted sampling effectively helps the model learn such minor features by detecting and emphasizing underrepresented samples. 

\subsection{Quantitative comparison of generated samples}

To verify that our method encourages model to generate underrepresented samples (having high discrepancy scores) for CIFAR-10 and CelebA, we evaluate `PFID (Partial FID)'. Original FID is calculated by comparing the feature statistics of all training data and randomly sampled generated samples, but PFID is calculated by the difference between the feature statistics of the specific subset of training data and generated samples. 
We evaluate the High PFID, the PFID of 5,000 training samples having the highest discrepancy scores and the Low PFID, the PFID of 5,000 training samples having the lowest discrepancy scores.
In both PFID calculations, we use 50,000 generated samples.
The results are summarized in Table \ref{table:fid_with_index} (averaged over three trials), where the PFID values are calculated for SNGAN.

This result shows the effectiveness of our method in two aspects. 
First, the Low PFID is significantly lower than the High PFID, which means that our discrepancy score successfully detects underrepresented samples. 
Another aspect is that after weighted sampling, the High PFID decreases significantly, implying the effectiveness of our method on promoting the consideration of high-scoring, underrepresented samples in GAN training.

\begin{table}[!ht]
    \caption{Partial FID for SNGAN}
    \label{table:fid_with_index}
    \centering
    \small
    \begin{tabular}{c|c|c|c}
    \toprule
    \multicolumn{2}{c|}{} & Baseline & Dia-GAN (ours) \\
    \midrule
    \multirow{2}{*}{CIFAR-10} & High PFID & {94.64}\scriptsize{$\pm$2.93} & 77.28\scriptsize{$\pm$3.76}\\
    \cmidrule{2-2} \cmidrule{3-4}
    & Low PFID & 22.43\scriptsize{$\pm$0.68} & 33.98\scriptsize{$\pm$1.77}\\
    \midrule
    \multirow{2}{*}{CelebA} & High PFID & 50.25\scriptsize{$\pm$3.24} & 42.33\scriptsize{$\pm$3.36}\\
    \cmidrule{2-2} \cmidrule{3-4}
    & Low PFID & 17.25\scriptsize{$\pm$1.35} & 23.17\scriptsize{$\pm$3.29}\\
    \bottomrule
    \end{tabular}

\end{table}

\section{Effect of our method in capturing semantic features for FFHQ}\label{sec:FFHQ}
To ensure that the ability of our method in capturing semantic features also applies to high-resolution datasets, we consider the FFHQ dataset and classify the race on the FFHQ dataset using the DeepFace architecture~\cite{deepface}. This architecture classifies the images as Asian, Black, Indian, Latino hispanic, Middle eastern, and White. For the FFHQ dataset, Black, Indian, and Middle eastern represent the minority taking less than 5\% of the FFHQ dataset. See Table \ref{tab:FFHQ_race} for details.
\begin{table}[!tb]
    \centering
    \caption{Ratio(\%) of race on the FFHQ dataset classified by the DeepFace architecture~\cite{deepface}.}
    \small
    \begin{tabular}{c|cccccc}
    \toprule
         Race & Asian & Black & Indian & Latino hispanic & Middle eastern & White \\
         \midrule
         Ratio(\%) & 19.38 & 4.80 & 2.08 & 10.83 & 4.04 & 58.87 \\
    \bottomrule
    \end{tabular}
    \label{tab:FFHQ_race}
\end{table}

We compare the occurrence rate and partial recall for these minor races after training with vanilla StyleGAN2 and Dia-GAN, respectively. The results are shown in Table \ref{table:FFHQ_occ_pr}.
\begin{table}[!tb]
\centering
\caption{FFHQ minor attribute analysis. O stands for the occurrence of minor attributes among the generated samples in percentage (\%) and R stands for the Partial Recall.}
\label{table:FFHQ_occ_pr}
\vspace{0.2em}
\small
\begin{tabular}{c|cc|cc}
\toprule
\multirow{2}{*}{} &\multicolumn{2}{c|}{Vanilla}&\multicolumn{2}{c}{\textbf{Dia-GAN}}\\
\cmidrule{2-5}
& O $\uparrow$ & R $\uparrow$ & O $\uparrow$ & R $\uparrow$ \\
\midrule
Black (4.80\%) & \textbf{3.00\scriptsize{$\pm$0.13}} & 0.27\scriptsize{$\pm$0.03} & 2.99\scriptsize{$\pm$0.17} & \textbf{0.30\scriptsize{$\pm$0.01}} \\
Indian (2.08\%) & 0.81\scriptsize{$\pm$0.19} & 0.26\scriptsize{$\pm$0.03} & \textbf{1.16\scriptsize{$\pm$0.03}} & \textbf{0.30\scriptsize{$\pm$0.01}} \\
Middle eastern (4.04\%) & 3.18\scriptsize{$\pm$0.20} & 0.27\scriptsize{$\pm$0.04} & \textbf{3.49\scriptsize{$\pm$0.61}} & \textbf{0.31\scriptsize{$\pm$0.00}} \\
\bottomrule
\end{tabular}
\end{table}

Similar to the results for the CelebA dataset in Section \ref{sec:exp_art}, the occurrence rate and partial recall for minor races in the FFHQ dataset are improved with our method, especially for Indian and Middle-eastern image samples.

In conclusion, this evidence demonstrates that our method successfully captures semantically meaningful minor attributes and emphasizes them during the training, resulting in a diverse generation of minor samples across low- to high-resolution datasets.

\clearpage
\newpage

\section{Examples of generated samples for MNIST-FMNIST}\label{supp:color_GAN}
\begin{figure*}[htb!]
\centering
\begin{tabular}{cccc}       
    & 99\% & 95\% & 90\% \\ [-1.8ex]
    \raisebox{0.09\linewidth}{Baseline} & \subfloat{\includegraphics[width=0.20\linewidth]{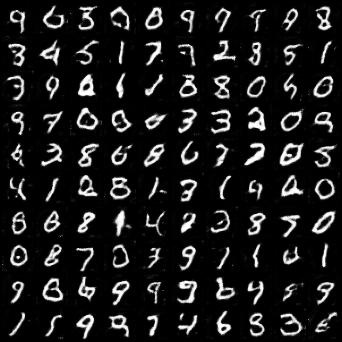}} &
    \subfloat{\includegraphics[width=0.20\linewidth]{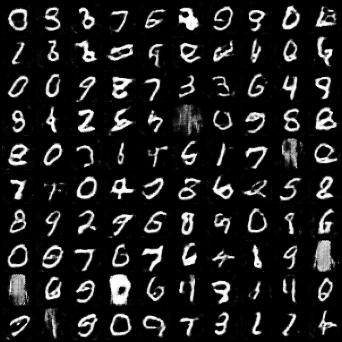}} &
    \subfloat{\includegraphics[width=0.20\linewidth]{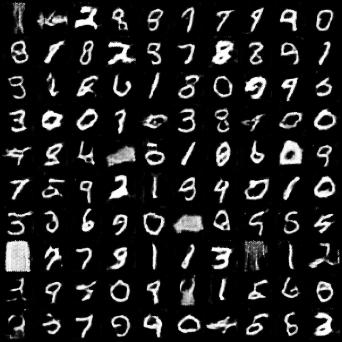}} \\ [-2.1ex]
    \raisebox{0.09\linewidth}{GOLD} & \subfloat{\includegraphics[width=0.20\linewidth]{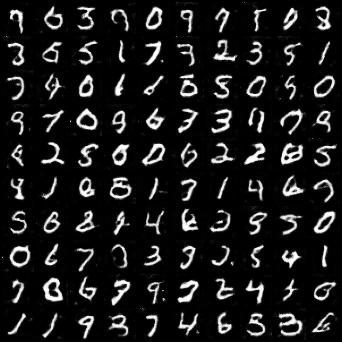}} &
    \subfloat{\includegraphics[width=0.20\linewidth]{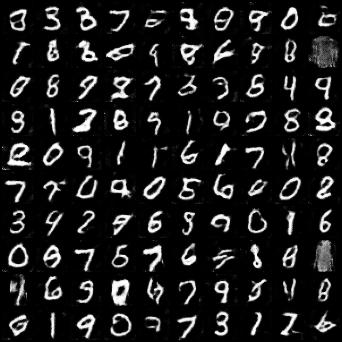}} &
    \subfloat{\includegraphics[width=0.20\linewidth]{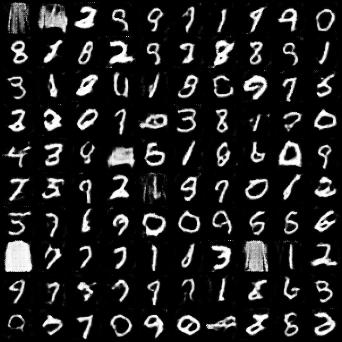}} \\ [-2.1ex]
    \raisebox{0.09\linewidth}{Top-k} & \subfloat{\includegraphics[width=0.20\linewidth]{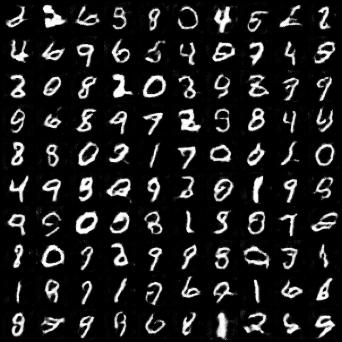}} &
    \subfloat{\includegraphics[width=0.20\linewidth]{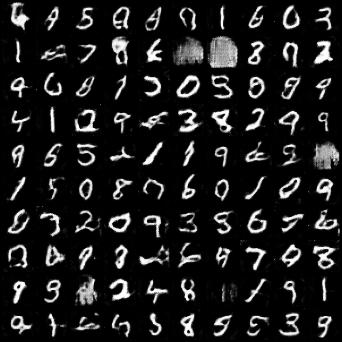}} &
    \subfloat{\includegraphics[width=0.20\linewidth]{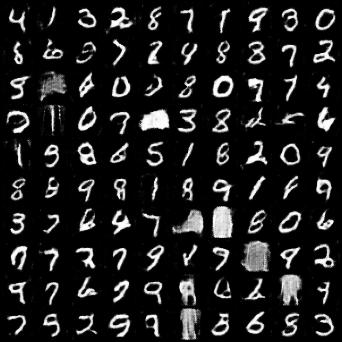}}\\ [-2.1ex]
    \raisebox{0.09\linewidth}{PacGAN} & \subfloat{\includegraphics[width=0.20\linewidth]{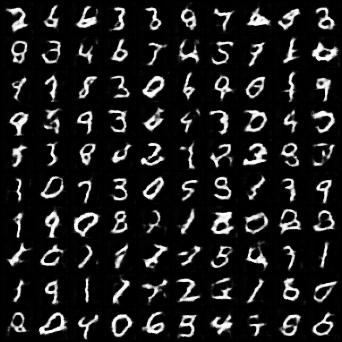}} &
    \subfloat{\includegraphics[width=0.20\linewidth]{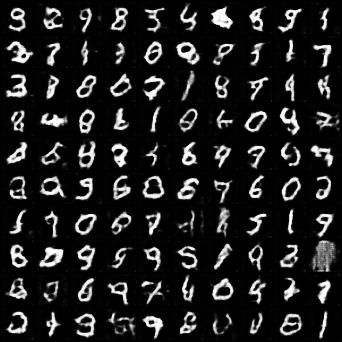}} &
    \subfloat{\includegraphics[width=0.20\linewidth]{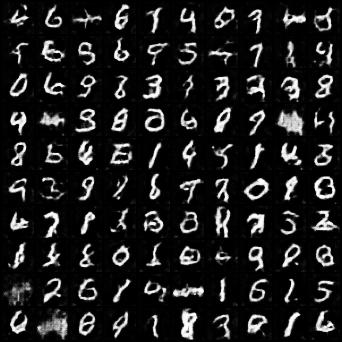}} \\ [-2.1ex]
    \raisebox{0.09\linewidth}{Inclusive GAN} & \subfloat{\includegraphics[width=0.20\linewidth]{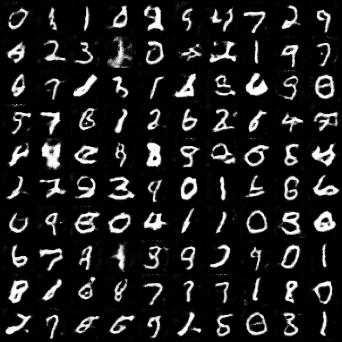}} &
    \subfloat{\includegraphics[width=0.20\linewidth]{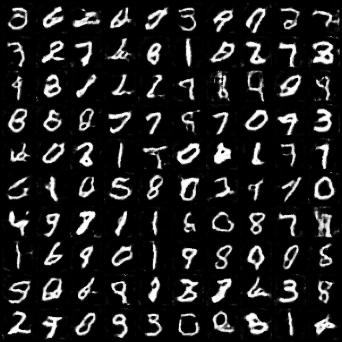}} &
    \subfloat{\includegraphics[width=0.20\linewidth]{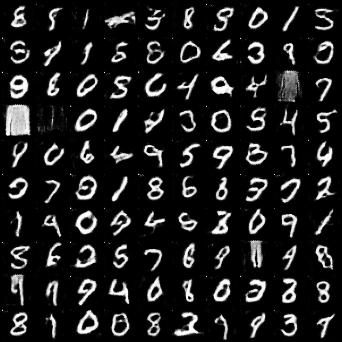}} \\ [-2.1ex]
    \raisebox{0.09\linewidth}{Dia-GAN} & \subfloat{\includegraphics[width=0.20\linewidth]{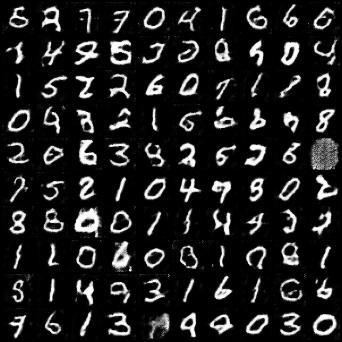}} &
    \subfloat{\includegraphics[width=0.20\linewidth]{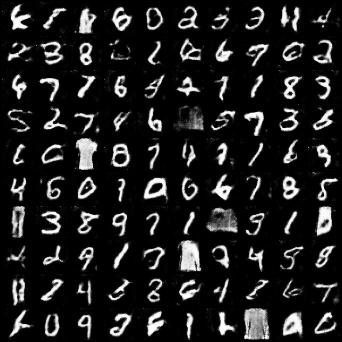}} &
    \subfloat{\includegraphics[width=0.20\linewidth]{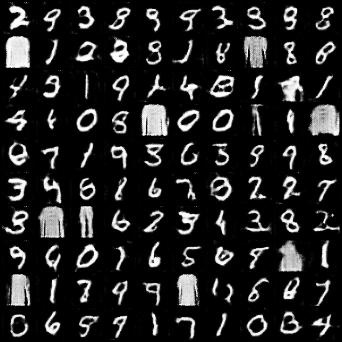}}
\end{tabular}
\caption{MNIST-FMNIST generated samples of various GANs on different majority rate.}
\label{fig:mnist-fmnist-supp}
\end{figure*}

We show randomly generated samples of various GANs trained on MNIST-FMNIST with different majority (MNIST) rate $\rho\in\{90,95,99\}\%$ in Fig.~\ref{fig:mnist-fmnist-supp}. 
Our method is the only method that recovers the minor (FMNIST) features when the rate is 99\%. 
Moreover, the reconstruction error (RE) scores reported in Table \ref{table:controlled_recon} demonstrate that our method improves the quality of generated samples with minor features, resulting in better RE score of the green training samples.
Results indicate the effectiveness of Dia-GAN in improving the quality of generated samples with underrepresented features.

\section{Full experimental results with standard deviation}\label{supp:full_result}
In this section, we show the detailed results for the tables in the main document, which are reported with only mean values due to the space limitation, with the standard deviation.

\begin{table}[!hbt]
\centering
\caption{(Details of Table~\ref{tab:minor_ldrv})
LDRV of major/minor groups on various datasets with majority rate 90\%.}
\label{tab:minor_ldrv_appendix}
\vspace{0.2em}
\begin{tabular}{c|ccc}
\toprule
Group &Gaussian ($\sigma$=3.0)  & Colored MNIST & MNIST-FMNIST \\
\midrule
Major & 0.001\scriptsize{$\pm$0.000} & 0.077\scriptsize{$\pm$0.018} & 0.082\scriptsize{$\pm$0.022} \\
Minor & 0.098\scriptsize{$\pm$0.009} & 0.186\scriptsize{$\pm$0.057} & 0.115\scriptsize{$\pm$0.021} \\
\bottomrule
\end{tabular}
\end{table}
\begin{table}[h]
\centering
\caption{(Details of Table~\ref{table:cifar10-celeba-results}) Comparison of diverse sampling/weighting techniques for CIFAR-10 image generation.}
\label{table:cifar10-fid_is_arxiv_v2}
\vspace{1em}
\begin{tabular}{c|cc|cc}
\toprule
\multirow{2.5}{*}{Methods} &\multicolumn{2}{c|}{SNGAN} &\multicolumn{2}{c}{SSGAN}\\
\cmidrule{2-5}
& \textbf{FID $\downarrow$}    & \textbf{IS $\uparrow$} & \textbf{FID $\downarrow$} & \textbf{IS $\uparrow$}\\
\midrule
Vanilla & 26.90\scriptsize{$\pm$0.90} & 7.36\scriptsize{$\pm$0.08} & 22.01\scriptsize{$\pm$0.99} & 7.65\scriptsize{$\pm$0.10}\\
DRS~\cite{DRS} & 24.54\scriptsize{$\pm$0.80} & 7.57\scriptsize{$\pm$0.05} & 20.51\scriptsize{$\pm$1.01} & 7.77\scriptsize{$\pm$0.09} \\
GOLD~\cite{GOLD} & 28.86\scriptsize{$\pm$0.92} & 7.21\scriptsize{$\pm$0.09} & 21.90\scriptsize{$\pm$0.90} & 7.57\scriptsize{$\pm$0.09} \\
GOLD + DRS~\cite{DRS} & 24.65\scriptsize{$\pm$0.86} & 7.53\scriptsize{$\pm$0.09} & 19.36\scriptsize{$\pm$0.45} & 7.79\scriptsize{$\pm$0.04} \\
Top-k~\cite{sinha2020top} &  24.45\scriptsize{$\pm$0.60} &  7.60\scriptsize{$\pm$0.06} & 20.01\scriptsize{$\pm$1.23} & 7.78\scriptsize{$\pm$0.08} \\
Top-k + DRS~\cite{DRS} &  23.92\scriptsize{$\pm$0.69} &  7.70\scriptsize{$\pm$0.09} & 20.09\scriptsize{$\pm$0.98} & 7.88\scriptsize{$\pm$0.10} \\
\textbf{Dia-GAN} & \textbf{19.66\scriptsize{$\pm$0.42}} & \textbf{7.95\scriptsize{$\pm$0.09}} & \textbf{16.31\scriptsize{$\pm$0.53}} & \textbf{8.14\scriptsize{$\pm$0.06}} \\
\bottomrule
\end{tabular}
\end{table}

\begin{table}[h]
\centering
\caption{(Details of Table~\ref{table:cifar10-celeba-results}) Comparison of diverse sampling/weighting techniques for CelebA image generation.} 
\label{table:celeba-fid_pr_arxiv}
\vspace{1em}
\begin{tabular}{c|ccc|ccc}
\toprule
\multirow{2.5}{*}{Methods} &\multicolumn{3}{c|}{SNGAN}&\multicolumn{3}{c}{SSGAN}\\
\cmidrule{2-7}
& \textbf{FID $\downarrow$}  & \textbf{Prec. $\uparrow$} & \textbf{Rec. $\uparrow$} & \textbf{FID $\downarrow$}  & \textbf{Prec. $\uparrow$} & \textbf{Rec. $\uparrow$}\\
\midrule
Vanilla & 7.12\scriptsize{$\pm$0.27} & 0.68\scriptsize{$\pm$0.00}  & 0.44\scriptsize{$\pm$0.01} & 7.19\scriptsize{$\pm$0.18} & 0.68\scriptsize{$\pm$0.01} & 0.44\scriptsize{$\pm$0.02} \\
DRS~\cite{DRS} & 7.04\scriptsize{$\pm$0.31} & 0.68\scriptsize{$\pm$0.01} & 0.44\scriptsize{$\pm$0.01} & 7.08\scriptsize{$\pm$0.23} & 0.68\scriptsize{$\pm$0.01} & 0.45\scriptsize{$\pm$0.01}\\
GOLD~\cite{GOLD} & 7.31\scriptsize{$\pm$0.67} & \textbf{0.69\scriptsize{$\pm$0.00}} & 0.44\scriptsize{$\pm$0.02} & 7.46\scriptsize{$\pm$0.31} & 0.68\scriptsize{$\pm$0.00} & 0.43\scriptsize{$\pm$0.00}\\
GOLD + DRS~\cite{DRS} & 6.97\scriptsize{$\pm$0.64} & 0.68\scriptsize{$\pm$0.01} & 0.44\scriptsize{$\pm$0.01} & 7.15\scriptsize{$\pm$0.37} & 0.67\scriptsize{$\pm$0.01} & 0.45\scriptsize{$\pm$0.01}\\
Top-k~\cite{sinha2020top} & 7.35\scriptsize{$\pm$0.20} & 0.67\scriptsize{$\pm$0.00} & 0.44\scriptsize{$\pm$0.01} & 7.23\scriptsize{$\pm$0.18} & 0.67\scriptsize{$\pm$0.00} & 0.45\scriptsize{$\pm$0.01}\\
Top-k + DRS~\cite{DRS} & 7.35\scriptsize{$\pm$0.18} & 0.68\scriptsize{$\pm$0.00} & 0.44\scriptsize{$\pm$0.00} & 7.16\scriptsize{$\pm$0.25} & \textbf{0.68\scriptsize{$\pm$0.00}} & 0.45\scriptsize{$\pm$0.00}\\
\textbf{Dia-GAN} & \textbf{6.70\scriptsize{$\pm$0.69}} & 0.64\scriptsize{$\pm$0.02} & \textbf{0.48\scriptsize{$\pm$0.02}} & \textbf{6.88\scriptsize{$\pm$0.58}} & 0.66\scriptsize{$\pm$0.02} & \textbf{0.46\scriptsize{$\pm$0.02}}\\
\bottomrule
\end{tabular}
\end{table}

\begin{table}[h]
\caption{(Details of Table~\ref{tab:stylegan2}) StyleGAN2 on FFHQ 256x256.}
\label{tab:stylegan2_appendix}
\vspace{0.2em}
\centering
\begin{tabular}{c|ccc}
    \toprule
        & FID $\downarrow$ & P $\uparrow$   & R $\uparrow$ \\
        \midrule
        StyleGAN2 & 14.07\scriptsize{$\pm$3.07}   & \textbf{0.72\scriptsize{$\pm$0.02}}  & 0.27\scriptsize{$\pm$0.03} \\
        GOLD    & 15.53\scriptsize{$\pm$4.14} & 0.69\scriptsize{$\pm$0.00}  & 0.29\scriptsize{$\pm$0.02} \\
        \textbf{Dia-StyleGAN2}  & \textbf{11.89\scriptsize{$\pm$0.21}} & 0.69\scriptsize{$\pm$0.01}  & \textbf{0.30\scriptsize{$\pm$0.01}} \\
        \bottomrule
\end{tabular}
\end{table}

\begin{table}[t]
\caption{(Details of Table~\ref{table:hingegan}) HingeGAN on CIFAR-10 and CelebA.}
\label{table:hingegan_appendix}
\vspace{0.2em}
\centering
\begin{tabular}{c|cc|c}
\toprule
\multirow{2}{*}{}     &  \multicolumn{2}{c|}{CIFAR-10} & CelebA\\
\cmidrule{2-4}
& FID $\downarrow$ & IS $\uparrow$ & FID $\downarrow$\\
\midrule
HingeGAN & 21.99\scriptsize{$\pm$1.73} & 7.67\scriptsize{$\pm$0.16} & 6.66\scriptsize{$\pm$0.06}\\
\textbf{Dia-HingeGAN} & \textbf{18.74\scriptsize{$\pm$1.79} }   & \textbf{8.02\scriptsize{$\pm$0.14}} & \textbf{5.98\scriptsize{$\pm$0.15}}\\
\bottomrule
\end{tabular}
\end{table}
\begin{table}[!tb]
\centering
\caption{(Details of Table~\ref{table:controlled_recon}) Reconstruction Error (RE) score of green (minor) training samples in Colored MNIST on different majority rate $\rho$.}
\label{table:controlled_recon_appendix}
\vspace{0.2em}
\begin{tabular}{c|ccc}
\toprule
Dataset & \multicolumn{3}{c}{Colored MNIST}\\
\midrule
Majority rate $\rho$ &{99\%} &{95\%} &{90\%}\\
\midrule
Vanilla & 0.838\scriptsize{$\pm$0.033}  & 0.236\scriptsize{$\pm$0.037}  & 0.218\scriptsize{$\pm$0.058}\\
GOLD~\cite{GOLD}  & 0.813\scriptsize{$\pm$0.002}& 0.297\scriptsize{$\pm$0.146}& 0.200\scriptsize{$\pm$0.022}\\
Top-k~\cite{sinha2020top}  & 0.831\scriptsize{$\pm$0.022}& 0.210\scriptsize{$\pm$0.012} & 0.223\scriptsize{$\pm$0.015}\\
PacGAN~\cite{lin2020pacgan}  & 0.810\scriptsize{$\pm$0.001}  & 0.244\scriptsize{$\pm$0.049} & 0.233\scriptsize{$\pm$0.052}\\
Inclusive GAN~\cite{InclusiveGAN} & 0.812\scriptsize{$\pm$0.001} & 0.274\scriptsize{$\pm$0.060}& 0.216\scriptsize{$\pm$0.024}\\
\textbf{Dia-GAN}  & \textbf{0.224\scriptsize{$\pm$0.020}}&  \textbf{0.204\scriptsize{$\pm$0.018}} & \textbf{0.197\scriptsize{$\pm$0.026}}\\
\bottomrule
\end{tabular}
\end{table}
\begin{table}[!tb]
\centering
\caption{(Details of Table~\ref{table:controlled_recon}) Reconstruction Error (RE) score of FMNIST samples (minor) in a mixture of MNIST and FMNIST on different majority rate $\rho$.}
\label{table:controlled_recon_appendix}
\vspace{0.2em}
\begin{tabular}{c|ccc}
\toprule
Dataset & \multicolumn{3}{c}{MNIST-FMNIST}\\
\midrule
Majority rate $\rho$ &{99\%} &{95\%} &{90\%} \\
\midrule
Vanilla & 0.290\scriptsize{$\pm$0.019}& 0.227\scriptsize{$\pm$0.001}& 0.215\scriptsize{$\pm$0.010}\\
GOLD~\cite{GOLD}  & 0.296\scriptsize{$\pm$0.008} & 0.241\scriptsize{$\pm$0.005} & 0.218\scriptsize{$\pm$0.004}\\
Top-k~\cite{sinha2020top}  & 0.281\scriptsize{$\pm$0.012} & 0.232\scriptsize{$\pm$0.006}& 0.221\scriptsize{$\pm$0.007}\\
PacGAN~\cite{lin2020pacgan}  & 0.313\scriptsize{$\pm$0.026} & 0.251\scriptsize{$\pm$0.013}& 0.225\scriptsize{$\pm$0.007}\\
Inclusive GAN~\cite{InclusiveGAN} & 0.283\scriptsize{$\pm$0.012} & 0.230\scriptsize{$\pm$0.015} & 0.220\scriptsize{$\pm$0.011}\\
\textbf{Dia-GAN} & \textbf{0.264\scriptsize{$\pm$0.007}} & \textbf{0.219\scriptsize{$\pm$0.016}} & \textbf{0.206\scriptsize{$\pm$0.002}}\\
\bottomrule
\end{tabular}
\end{table}
\begin{table}[!tb]
\centering
\caption{(Details on Table~\ref{table:celeba_attr_count_ldrv}) CelebA minor attribute analysis. Mean of LDRV and mean of the discrepancy score of CelebA samples with (W/) or without (W/O) minor attributes.}
\label{table:celeba_attr_ldrv_appendix}
\vspace{0.2em}
\begin{tabular}{c|cc|cc}
\toprule
\multirow{2.5}{*}{Method}&\multicolumn{2}{c|}{LDRV}&\multicolumn{2}{c}{Discrepancy}\\
\cmidrule{2-5}
& W/ & W/O & W/ & W/O\\
\midrule
Bald (2.244\%) & \textbf{0.271\scriptsize{$\pm$0.050}} & 0.184\scriptsize{$\pm$0.035} & \textbf{2.938\scriptsize{$\pm$0.183}} & 2.221\scriptsize{$\pm$0.183} \\
Double Chin (4.669\%) & \textbf{0.219\scriptsize{$\pm$0.040}} & 0.184\scriptsize{$\pm$0.035} & \textbf{2.525\scriptsize{$\pm$0.188}} & 2.224\scriptsize{$\pm$0.183}\\
Eyeglasses (6.512\%) & \textbf{0.254\scriptsize{$\pm$0.048}} & 0.181\scriptsize{$\pm$0.035} & \textbf{2.783\scriptsize{$\pm$0.202}} & 2.200\scriptsize{$\pm$0.182}\\
Gray Hair (4.195\%) & \textbf{0.211\scriptsize{$\pm$0.037}} & 0.185\scriptsize{$\pm$0.035} & \textbf{2.450\scriptsize{$\pm$0.173}} & 2.228\scriptsize{$\pm$0.184}\\
Mustache (4.155\%) & \textbf{0.242\scriptsize{$\pm$0.047}} & 0.183\scriptsize{$\pm$0.035} & \textbf{2.699\scriptsize{$\pm$0.218}} & 2.218\scriptsize{$\pm$0.182}\\
Pale Skin (4.295\%) & \textbf{0.190\scriptsize{$\pm$0.032}} & 0.186\scriptsize{$\pm$0.036} & \textbf{2.240\scriptsize{$\pm$0.156}} & 2.238\scriptsize{$\pm$0.184}\\
Wearing Hat (4.846\%) & \textbf{0.357\scriptsize{$\pm$0.072}} & 0.177\scriptsize{$\pm$0.034} & \textbf{3.651\scriptsize{$\pm$0.297}} & 2.164\scriptsize{$\pm$0.178}\\
\bottomrule
\end{tabular}
\end{table}
\begin{table}[!tb]
\centering
\caption{(Details of Table~\ref{table:celeba_attr_count_ldrv}) CelebA minor attribute analysis. O stands for the occurrence of minor attributes among the generated samples in percentage (\%) and R stands for the Partial Recall.}
\label{table:celeba_attr_count_appendix}
\vspace{0.2em}
\begin{tabular}{c|cc|cc}
\toprule
\multirow{2.5}{*}{Method} & \multicolumn{2}{c|}{Vanilla} &\multicolumn{2}{c}{\textbf{Dia-GAN}}\\
\cmidrule{2-5}
& O $\uparrow$ & R $\uparrow$ & O $\uparrow$ & R $\uparrow$ \\
\midrule
Bald (2.244\%) & 0.678\scriptsize{$\pm$0.164} & 0.353\scriptsize{$\pm$0.014} & \textbf{0.836\scriptsize{$\pm$0.089}} & \textbf{0.393\scriptsize{$\pm$0.012}} \\
Double Chin (4.669\%) & 0.440\scriptsize{$\pm$0.090} & 0.411\scriptsize{$\pm$0.015} & \textbf{0.522\scriptsize{$\pm$0.090}} & \textbf{0.461\scriptsize{$\pm$0.003}}\\
Eyeglasses (6.512\%) & 3.300\scriptsize{$\pm$0.044} & 0.400\scriptsize{$\pm$0.019} & \textbf{4.053\scriptsize{$\pm$0.282}} & \textbf{0.449\scriptsize{$\pm$0.008}}\\
Gray Hair (4.195\%) & 2.273\scriptsize{$\pm$0.335} & 0.402\scriptsize{$\pm$0.016} & \textbf{2.369\scriptsize{$\pm$0.087}} & \textbf{0.436\scriptsize{$\pm$0.013}}\\
Mustache (4.155\%) & 0.157\scriptsize{$\pm$0.027} & 0.391\scriptsize{$\pm$0.012} & \textbf{0.228\scriptsize{$\pm$0.009}} & \textbf{0.433\scriptsize{$\pm$0.008}}\\
Pale Skin (4.295\%) & 0.346\scriptsize{$\pm$0.014} & 0.380\scriptsize{$\pm$0.013} & \textbf{0.453\scriptsize{$\pm$0.004}} & \textbf{0.427\scriptsize{$\pm$0.025}}\\
Wearing Hat (4.846\%) & 2.307\scriptsize{$\pm$0.055} & 0.380\scriptsize{$\pm$0.007} & \textbf{3.595\scriptsize{$\pm$0.655}} & \textbf{0.408\scriptsize{$\pm$0.020}}\\
\bottomrule
\end{tabular}
\end{table}
\end{document}